\RequirePackage{fix-cm}
\documentclass[smallextended]{svjour3}       % onecolumn (second format)
\smartqed  % flush right qed marks, e.g. at end of proof
\usepackage{graphicx}
\usepackage{natbib}
\usepackage{hyperref}
\usepackage{xcolor}
\definecolor{darkblue}{rgb}{0, 0, 0.5}
\hypersetup{colorlinks=true,citecolor=darkblue, linkcolor=darkblue, urlcolor=darkblue}

\usepackage[utf8]{inputenc}

\usepackage{enumitem}

\begin{document}

\title{Towards transparency in NLP shared tasks
\thanks{The first author was supported by the European Union’s Horizon 2020 research and innovation programme under the Marie Skłodowska-Curie grant agreement N\textordmasculine~713567 and Science Foundation Ireland in the ADAPT Centre (Grant 13/RC/2106) (\href{www.adaptcentre.ie}{www.adaptcentre.ie}) at Dublin City University. The second author was funded by the Irish Government Department of Culture, Heritage and the Gaeltacht under the GaelTech Project at Dublin City University.}
}
% Grants or other notes about the article that should go on the front
% page should be placed within the \thanks{} command in the title
% (and the %-sign in front of \thanks{} should be deleted)
%
% General acknowledgments should be placed at the end of the article.



\author{Carla Parra Escartín         \and
        Teresa Lynn \and
        Joss Moorkens \and
        Jane Dunne
}


\institute{Carla Parra Escartín \at
              Iconic Translation Machines, Ltd. \\
              Dublin, Ireland \\
              \email{carla@iconictranslation.com}           %  \\
           \and
           Teresa Lynn \at
              ADAPT Centre, Dublin City University \\
              Dublin, Ireland \\
              \email{teresa.lynn@adaptcentre.ie}
           \and
           Joss Moorkens \at
              ADAPT Centre, Dublin City University \\
              Dublin, Ireland \\
              \email{joss.moorkens@adaptcentre.ie}
           \and
           Jane Dunne \at
              ADAPT Centre, Dublin City University \\
              Dublin, Ireland \\
              \email{jane.dunne@adaptcentre.ie}
}

\date{Received: date / Accepted: date}

\maketitle

\begin{abstract}
This article reports on a survey carried out across the Natural Language Processing (NLP) community. The survey aimed to capture the opinions of the research community on issues surrounding shared tasks, with respect to both participation and organisation. Amongst the 175 responses received, both positive and negative observations were made. We carried out and report on an extensive analysis of these responses, which leads us to propose a Shared Task Organisation Checklist that could support future participants and organisers.

The proposed Checklist is flexible enough to accommodate the wide diversity of shared tasks in our field and its goal is not to be prescriptive, but rather to serve as a tool that encourages shared task organisers to foreground ethical behaviour, beginning with the common issues that the 175 respondents deemed important. Its usage would not only serve as an instrument to reflect on important aspects of shared tasks, but would also promote increased transparency around them.
\keywords{Shared Tasks \and Transparency in NLP \and Survey \and Ethics in NLP}
\end{abstract}

\section{Introduction}
\label{sec:intro}

Shared tasks are community-organised challenges or competitions designed to allow teams of researchers to submit systems to solve specific tasks.\footnote{Note that depending on the area of research, different terms are used to refer to these competitions, such as challenges, benchmarks and evaluation campaigns. In this paper, we refer to all of them with the collective term ``shared task'' notwithstanding the area of research they belong to.} The current landscape of shared task activity in the field of Natural Language Processing (NLP) is wide and varied. \citet{parraescartin-EtAl:2017:EthNLP} reported an increase in the number of shared tasks being proposed at major NLP conferences in 2016. Alongside annual calls for specific shared tasks, new ones regularly emerge, 
%This tendency seems to have prevailed in our field,
with many conferences now including a shared task track among thematic workshops. This article reports on a survey carried out in 2017 relating to shared tasks in the field of 
%Natural Language Processing (
NLP.\footnote{The survey contained a short Plain Language Statement and an Informed Consent form. It was approved by the relevant Research Ethics Committee of the authors' affiliation.} The purpose of this survey was two-fold. Firstly, to get an overview of the variety of aspects relating to shared tasks in the field. Through this overview, we aimed to gain insight into what was considered to be best practice in shared task organisation and participation.

Secondly, through putting specific questions to the NLP community and providing the opportunity for additional feedback, we anticipated that the outcome of this survey would form a basis for the creation of a common framework for shared tasks in NLP. The aim here is not to criticise specific shared tasks, but rather to focus on the lessons learnt by our respondents, share them with the community in general and to learn together from our collective knowledge.%, and also to highlight areas for concern and take steps towards putting best practices in place.
%also make the learning curve to access shared tasks more accessible to newcomers to the field. 

%The current landscape of shared task activity is wide and varied. \citet{parraescartin-EtAl:2017:EthNLP} reported an increase in the number of shared tasks being proposed at major conferences in NLP in 2016. Alongside yearly calls for specific shared tasks, new ones emerged. This tendency seems to have prevailed in our field, with many conferences including now a shared task track in many of their thematic workshops. 
In order to make a reliable assessment and to propose ideas for best practices going forward, input from the broader NLP community was required (as opposed to relying on the relatively limited experiences or opinions of a small number of authors and anecdotal evidence from their colleagues or collaborators).
%For this reason, an online survey was
To that end, we created and shared an online survey across a number of different media (e.g. mailing lists, social media platforms, etc.\footnote{We used the Twitter handle @STaskSurvey and the hashtag \#STaskSurvey}). The survey covered 30 questions, most of which (27) were compulsory. Questions in the survey addressed two main perspectives -- that of \textit{participating} in and that of \textit{organising} a shared task. Additionally, we had three open questions at the end to allow participants to share particularly good or bad experiences, or to provide any other additional comments.

This article is organised as follows: Section~\ref{sec:motivation} explains the motivating factors behind this survey along with describing the goal of the study and how it was designed; Section~\ref{sec:surveyResults} reports on the results of the survey in two sub-sections: participation (Section~\ref{ssec:participationResults}) and organisation (Section~\ref{ssec:organisationResults}). Section~\ref{sec:discussion} provides a discussion of the survey results and our proposed common framework for shared tasks.

\section{Motivation and Background}
\label{sec:motivation}

Shared tasks in NLP are becoming increasingly common, and with their growth in numbers comes a wide variability in the manner in which they are organised. This includes major differences in aspects such as the requirements for participation, the nature of data management and the degrees of transparency required for reporting, for example. While there is bound to be variability across shared tasks -- with the varying nature of tasks, this will be inevitable -- there are certain universal factors that could apply to all or many.
Both \citet{parraescartin-EtAl:2017:EthNLP} and \citet{SharingIsCaring} acknowledge that organisation of shared tasks has evolved in both a haphazard and a ``by convention" manner in the absence of available guidelines or standards. And while we can assume general good behaviour, that tasks are organised in good faith, and that on the whole both organisers and participants behave ethically, an increase in transparency would benefit our community and allow future participants and organisers to follow established best practices.

\citet{parraescartin-EtAl:2017:EthNLP} brought the inconsistency and lack of framework for shared task organisation to the attention of the community in the context of ethical considerations. While the paper did not set out to assert that there are major issues at present, it highlighted that the organic nature of shared task formation brings with it potential risks for the future. Their overview of these risks demonstrated the need for a common framework that could mitigate against potential future issues by ensuring transparency.

\subsection{Potential issues related to shared tasks}

%Shared tasks are community-organised challenges or competitions designed to allow teams of researchers to submit systems to solve specific tasks. 
In the field of NLP, these predefined tasks usually reflect contemporary technical challenges for the research community. Therefore, the main purpose of these shared tasks is to encourage researchers to combine their skills and strengths in an effort to ``solve" or address the problem at hand. The research community benefits from advances in state-of-the-art approaches that emerge as a result, new data sets and standards are often developed, creativity and motivation amongst researchers is fostered, and the winners receive international recognition. Often this is achieved through a leaderboard culture, whereby models or systems are ranked using performance-based evaluation on a shared benchmark. And while this is generally accepted, it has also been criticised, as these leaderboards many not always reflect the best model for a real-world application in their ranking, where other factors such as latency or prediction cost may have to be weighted in together with model accuracy \citep{Ethayarajh2020}. 
%Often this is achieved through a leaderboard culture, whereby models or systems are ranked using performance-based evaluation on a shared benchmark \citep{Ethayarajh2020}. 

While shared tasks are not a new phenomenon (some date back as far as the 1980s\footnote{See \citet{Pallett:2003} for an overview of these first shared tasks.}), there has been an increase in the number of shared tasks over the past few years. \citet{SharingIsCaring} calculate that over 100 shared tasks took place between 2014-2016 and \citet{parraescartin-EtAl:2017:EthNLP} note that ACL2016 alone included an additional 9 shared tasks when compared with the previous year's event. This tendency seems to have continued since the publication of those findings, and shared tasks are still growing up in our field. For example, in 2020, 67 different shared tasks were organised across 3 major NLP conferences: ACL 2020\footnote{\url{https://acl2020.org}} (20 shared tasks), EMNLP 2020\footnote{\url{https://2020.emnlp.org}} (20 shared tasks) and COLING 2020\footnote{\url{https://coling2020.org}} (27 shared tasks). This is not surprising, as co-locating a workshop featuring a shared task with an established conference that addresses a similar topic can help to attract engagement from within the community. This can not only become an opportunity to gain momentum on a particular research topic, but can also translate into a larger number of participants for that particular workshop if the shared task gains sufficient attention. What is also unsurprising, given the lack of a framework or guidelines for running shared tasks, is the range of variation in their set-up. \citet{parraescartin-EtAl:2017:EthNLP} highlight potential risks that accompany such a haphazard approach. Here we provide a brief summary of those potential issues (PIs): 

\begin{itemize}
\item \textbf{(PI-1) Secretiveness}\footnote{Fortunately, in the last few years a trend has emerged in our field, whereby reviewers to major conferences are also asked to evaluate to which extent the results published in a paper are reproducible and the code and/or additional data is shared. This has fostered a situation whereby an increasing amount of authors now release the data and/or code they have used and more details about the research carried out.}

In competitive environments, there can sometimes be an element of secretiveness with respect to sharing the details of a system.

While participants are often required to submit a written description of their system to the shared task organiser, the way in which they are described or the level of detail provided may vary.

\item \textbf{(PI-2) Unconscious overlooking of ethical considerations}

Research teams and institutions tend to consider ethics only for studies that involve humans, animals, or personal data, whereas ethics may not be at the forefront of the minds of those organising or participating in a shared task event. As an example related to PI-1, secretive behaviour is not always intentional, and vagueness in terms of reporting often arises from convention (i.e. other teams have previously reported their system with minimal details). Lack of guidelines or standards are likely to lead to such reporting practices.

\item \textbf{(PI-3) Redundancy and replicability in the field}\footnote{While we have observed a trend in recent years (c.f. previous footnote), replicability is still an issue in our field. Even though more authors now share their experimental and data setup, very few studies actually test the replicability of research papers. A recent effort in this regard is the REPROLANG2020 workshop at LREC2020, which was organised around a shared task aimed at fostering reproducibility \citep{branco-EtAl:2020:LREC2}.}

Arising from issues such as vaguely-detailed reporting is the issue of lack of replicability. If insufficient information on system development is shared, or if additional data is not released, it becomes difficult for others to replicate a study. This can be particularly challenging if other researchers want to use that system as a baseline in future experiments. 
The need for reproducibility and replicability is a widely-discussed topic in the field as a whole and its importance has been well documented by \citet{ReplicArticle-Branco2017}.

\item \textbf{(PI-4) Potential conflicts of interest}

In many shared tasks, there is a clear delineation between those who organise and those who participate. There is an understanding that organisers have privileged access to test sets, or an in-depth knowledge of the data that may not be known to participants, which could provide an unfair advantage. However, there are always scenarios where there is overlap with the two groups, and organisers do in fact participate in the competition.  While it is not implied here that bad practices occur, we believe that encouraging a higher degree of transparency would be beneficial for our community.

\item \textbf{(PI-5) Lack of description of negative results}

A common issue in shared tasks is the inclination for participants to report only positive results from their systems, possibly influenced by the leaderboard culture. At the same time, some shared tasks do not in fact insist on full system reporting. We contend that this lack of reporting is not done consciously, but rather because researchers tend to be more prone to publishing positive results. The drawback from such practice is that the value of learning through negative results is lost (i.e. knowledge of which approaches do not work well).

\item \textbf{(PI-6) Withdrawal from competition}

Closely linked to a reluctance towards reporting negative results is the practice of withdrawing from a competition. Driven by the fear of reporting low results or ranking poorly, this sometimes occurs when participants realise that their system is not doing well in the lead-up to the submission deadline.\footnote{Of course, other issues such as technical problems or running out of time due to various factors may also cause withdrawals.}

\item \textbf{(PI-7) Potential `gaming the system'}

Shared tasks use evaluation metrics to rank the quality of submitted systems. However, in some scenarios, tuning only to automatic metrics can mean losing sight of improving real-world applications. For example, in machine translation, the BLEU metric (which has become \textit{de facto} standard) is often criticised as being a misleading indicator of translation quality in terms of fluency and adequacy. And yet, despite the criticism, many researchers would acknowledge that when training their systems they focus on tuning towards that specific metric.

\item \textbf{(PI-8) Unequal playing field}

As shared tasks are generally open to anyone, there can be widely-varying types of participants -- from a single researcher in a small low-funded research lab to a large industry-based team who has access to the best resources on offer. Resources could include not only processing power and larger teams but also language resources and tools (e.g. corpora, pre-processing tools).
In some shared tasks scenarios, this can be viewed as an unequal playing field.

\end{itemize}

\section{The Shared Tasks Survey}
\label{sec:surveyResults}

\subsection{The Survey and Questionnaire Design}
\label{ssec:surveyDesign}

The survey was launched on April 4th 2017 and continued until June 30th 2017, and was shared across mailing lists and social media platforms. The survey was anonymous and the results have been held on a secure server. It comprised 30 questions, most of which (27) were compulsory. In addition, most questions provided a comment field, allowing for (optional) additional feedback from the respondents. The survey utilised snowball sampling \cite{Johnson_snowball} in order to maximise the number of respondents and was disseminated online in order to gather honest opinions \cite{SussmanSproull}.\footnote{\citet{SussmanSproull} show that people are likely to be more honest over online communication, particularly when they need to deliver bad news. Thus, an online survey would potentially allow respondents to share negative experiences or criticism more openly than if we had opted for a focus group or interviews.}

One goal of the survey was to receive input and feedback on the topic of shared tasks from as broad a range of disciplines as possible in the area of language technology. As will be seen in this section, we received responses from researchers across a diverse range of disciplines, all of which have varying types of shared tasks. We use these statistics relating to shared task participation as metadata to better interpret some responses throughout the survey. 

In designing the survey and questionnaire we endeavoured to follow best practices as described in \citet{lazarEtAl2010HC}. 176 people started our survey, with one respondent declining to confirm consent to the terms of participation in the first step. Thus in total, we had 175 respondents to the first question in the survey. Given that the survey was announced and promoted through social media, we had no control over who was going to respond. Therefore, we devised mechanisms for qualifying potential respondents to our survey.\footnote{In order to avoid confusion, we choose ``respondent" to refer to those who completed our survey, as we are using the term ``participants" in reference to those involved in taking part in a shared task.} 
%The questionnaire began with a few profiling questions such as whether they knew what a shared task is, or whether they had participated in any.

The first question in the survey asked whether the respondent understood the concept of a shared task. Those who did not (eight respondents) were thanked for their time and not exposed to the subsequent questions. Therefore, the analysis and discussion below relate to the responses of \textbf{167 respondents}. \citet{SueRitter2007} report that a minimum of 30 responses should be considered as a baseline for survey research \cite{lazarEtAl2010HC}. As our number of questionnaire respondents is more than 5 times that figure, we deem the results summarised in this article to be valid. 

Additionally, according to  \citet{lazarEtAl2010HC}, both \citet{McKennaBargh} and \citet{SpearsLea} point out that anonymous surveys ``may also lead to an increased level of self-disclosure". As such, our respondents were never asked for their identity.

The survey was divided into two parts, relating to (i) participation in shared tasks (Section~\ref{ssec:participationResults}) and (ii) organisation of shared tasks (Section~\ref{ssec:organisationResults}). This structure was chosen in order to account for the different experiences of shared task participants and organisers. While some aspects are shared, the perspective from which a respondent answered the questions would allow for broader feedback.  %This structure was chosen in order to account for the fact  that the experiences of participants and organisers can greatly differ. 

Some questions gave respondents the opportunity to provide additional information through a comments field.  Comments were not compulsory, as we were aware that respondents may not have the time to add additional information to their responses or have strong enough feelings on a subject to warrant additional input. 
However, those comments that were offered by respondents provided insight into a number of factors impacting this study, including (i) the certainty/uncertainty of some of the responses, (ii) strong arguments \textit{for} or \textit{against} a particular feature of shared tasks, and (iii) the clarity (or lack thereof) of survey questions. In our analysis, we take comments into account as well as reporting on absolute responses. 

There are a couple of features of the design of this survey worth mentioning here. Firstly, as is often deemed more accurate in capturing questionnaire responses \cite{Likert-Nadler}, we chose a Likert-type scale of 4 (forced choice) where the responses featured a degree of opinion between Never -- Always or Strongly Disagree -- Strongly Agree. A limitation we encountered with Google Forms meant that only the two extremes were labelled, but the middle ground responses were left to interpretation (e.g. sometimes, often, seldom, etc). Secondly, while the scale worked well for the most part, through the comments provided, we noted some instances where a ``Not Sure" option may have been appropriate (e.g. c.f ~\ref{sssec:organiserParticipation}, ~\ref{sssec:annotatorParticipation}).

According to \citet{lazarEtAl2010HC} for a survey to be valid, its questionnaire should be well written and questions should be as un-biased as possible. Therefore, when preparing the questionnaire special attention was paid to the types and structure of questions.

We first prepared a set of questions to be included in the questionnaire, and then ensured that each question only asked one question (i.e. ``double-barrelled'' questions were avoided). Where possible we kept the polarity of the questions in the affirmative case to avoid confusing respondents. Finally, we tried to phrase questions in a respondent-centric way, whereby they were directly asked for their experience or opinion.

A first draft of the questionnaire was reviewed by two independent reviewers to identify any ambiguous questions or needs to adjust the questionnaire structure. All necessary adjustments were made prior to launching the survey.

In the following sections we discuss the results of the shared tasks survey. We do this by stepping through each question asked in the survey (in order) and reporting on both the responses (with accompanying charts) and associated comments provided by respondents. We also highlight the relevant Potential Issue (PI) for each question (see Section~\ref{sec:motivation} for an explanation of these PIs). The comments provided in the free text boxes were analysed by the authors and labelled across recurrent themes. Rather than starting out with a pre-defined taxonomy, the labels were created alongside our analysis and tailored to each individual question.

\subsection{Questions relating to participation in NLP shared tasks}
\label{ssec:participationResults}

In response to our profiling questions, of the 167 respondents who knew what a shared task was, 133 of them had previously participated in a shared task, 9 respondents had not, and 5 respondents had considered it but had changed their minds (pulled out) (PI-6).

The reasons given for pulling out of shared tasks were:
%\begin{itemize}
(i) the onerous barrier for entrance; 
(ii) sparse documentation, datasets proving difficult to obtain, previous results being hard or impossible to reproduce, etc.;
(iii) an inability to cope with both the unexpected increase in the workload and the requirements of the shared task in question; (iv) the system was not ready in time; (v) too many other work commitments.
%\end{itemize}

\subsubsection{Frequency of participation}  \textcolor{white}{word}\\   
\label{sssec:paxFrequency}
%"\texttt{How many times have you participated in a shared task?}"

\begin{figure}[h!]
\begin{center}
\includegraphics[scale=0.55]{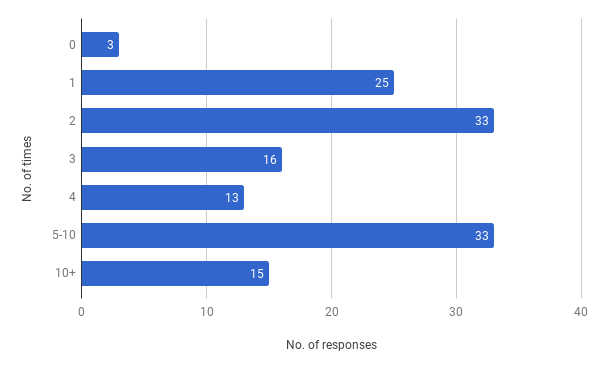}
\caption{"\texttt{How many times have you participated in a shared task?}".}
\label{fig:ST_Participation}
\end{center}
\end{figure}

In order to understand how informed the survey responses are, we asked the respondents how many times they participated in a shared task. Figure \ref{fig:ST_Participation} demonstrates the frequency of participation of the 138 respondents who participated in shared tasks to some degree.\footnote{Note that three of the five who pulled out of a shared task referred to that event as non-participation. The remaining two acknowledge their activities as participation. This discrepancy in responses is possibly due to the stage at which participation ended.}
80\% of respondents had participated in two or more shared tasks, with 34\% having participated in five or more shared tasks. This result indicates that the majority of the respondents have solid experience of shared task participation.
Note that we capture these statistics as metadata in presenting our findings in Figures~\ref{fig:OpinionTransparency} to ~\ref{fig:replicabilityPax2}.

\subsubsection{Cross-section of shared tasks}
\label{sssec:STselection}

%\texttt{"Please select all shared tasks in which you have participated"}

%CPE: I commented out the previous versions of the figures and created a new chart. We had 3 charts that showed the same from different angles.

%\begin{figure}[h!]
%\begin{center}
%  \includegraphics[scale=0.70]{Images/4_1_2a.png}
%\caption{\texttt{"Please select all shared tasks in which you have participated"}}
%\end{center}
%\end{figure}

%\begin{figure}[h!]
%\begin{center}
%\includegraphics[scale=0.70]{Images/4_1_2b.png}
%\caption{}
%\end{center}
%\end{figure}

%\begin{figure}[h!]
%\begin{center}
%\includegraphics[scale=0.70]{Images/4_1_2c.png}
%\caption{}
%\end{center}
%\end{figure}

In order to capture the extent of discipline variation that informed this survey, we requested information on the type of shared task participation. We provided a list of twelve well-known shared task options to choose from and asked respondents to name their relevant shared task if not listed.
In total, 138 respondents reported to have participated in as many as 33 different shared tasks. Combined, they accounted for 276 different shared task submissions.

\begin{figure}[h!]
\begin{center}
  \includegraphics[scale=0.55]{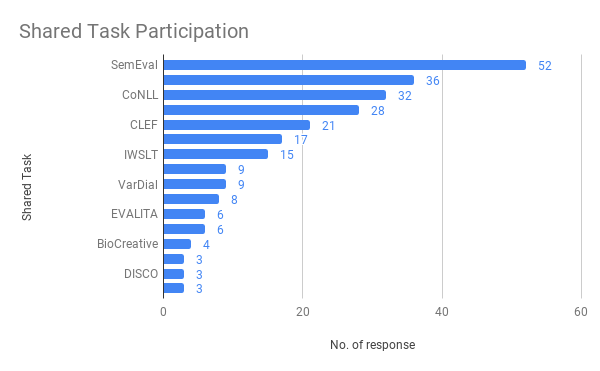}
\caption{\texttt{"Please select all shared tasks in which you have participated"}.}
\label{fig:FreqOfParticipation}
\end{center}
\end{figure}

The chart in Figure \ref{fig:FreqOfParticipation} shows the distribution across those shared tasks that obtained at least three or more hits for participation. The most represented shared tasks were SemEval (38\%), CoNLL (23\%), and WMT-related shared tasks (46\%, with 25\% of those WMT respondents indicating that they had participated in the WMT Translation shared tasks, and 20\% indicating participation in other WMT shared tasks).\footnote{These percentages do not amount to 100, as we are computing the number of respondents participating in each shared task. As this question was answered by means of check boxes, the respondents indicated \textit{all} shared tasks they had participated in (i.e. each respondent could select one or more shared tasks they had participated in, and even add new ones to the list).}
The ``Other" field allowed respondents to indicate shared tasks that were not listed. In total 58 additional shared tasks were indicated by our respondents. In some cases there were overlaps across responses: BEA, DISCO and SPMRL were indicated three times, whereas a further 13 shared tasks were indicated twice: *SEM, BioNLP, CLIN, CLPSYCH, EMPIRIST, NAACL, NLI, NTCIR, PASCAL, SANCL, TASS, TweetMT and WNUT. Forty-one shared tasks were mentioned only once.\footnote{In the interest of space, we are not listing these tasks here.} This wide variety of responses highlights the vast number of shared tasks that exist in the field.

\subsubsection{Opinion on transparency of shared task organisation (PI-4)}
\label{sssec:transparency}
%\vspace{0.2cm}
%\texttt{"I feel that the shared tasks I participated in were organised in a clear and transparent way"}\\

\begin{figure}[!ht]
\begin{center}
\includegraphics[scale=0.65]{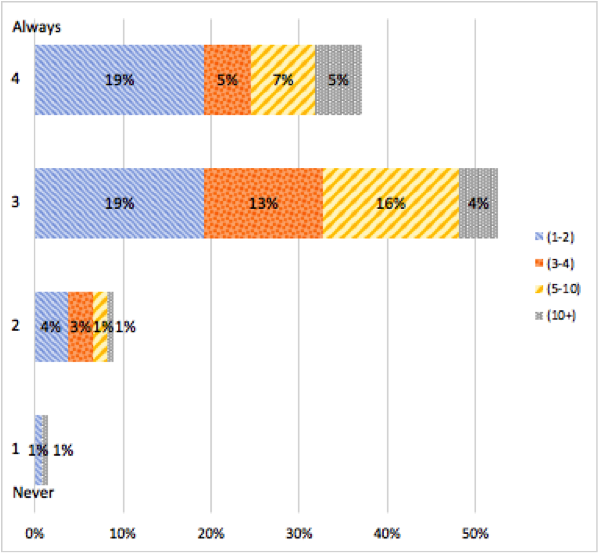}
\caption{\texttt{"I feel that the shared tasks I participated in were organised in a clear and transparent way"}}
\label{fig:OpinionTransparency}
\end{center}
\end{figure}

Figure \ref{fig:OpinionTransparency} shows the respondents' opinions with respect to the transparency of the organisation of shared tasks they had participated in. To provide further insights, it also indicates the overall number of shared tasks the respondents had previously participated in.\footnote{From this point on, we use in our analysis this combined approach of showing responses to a particular question in our survey combined with metadata extracted from the respondent's profiling questions. In this section, we use the metadata related to the total amount of shared tasks in which respondents participated.} The answers to this question were provided on a Never -- Always scale of 4.

As can be observed in Figure \ref{fig:OpinionTransparency}, overall there seems to be a general consensus among respondents that they were happy with the clarity and transparency surrounding the shared tasks in which they participated. Some accompanying comments noted that the answer depended on the task, i.e. they had positive experiences with some and negative experiences with others. Supporting comments to explain dissatisfaction in some instances varied from unclear guidelines, data formatting inconsistencies, unclear definitions of metrics, and multiple changes of datasets over the course of the competition. These results are particularly interesting, because as we shall see, they seem to contradict their own responses to the questions appearing later on in the questionnaire. One possible hypothesis to explain this change of opinion could be that this was one of the first questions in the questionnaire, and perhaps respondents had not fully reflected on their experience in the level of detail that we invited them to do with the subsequent questions. Further insight into this topic was received at the end of the questionnaire which is addressed in Section~\ref{ssec:finalremarks}.

%These comments are also in line with the comments of one respondent who selected \textit{Never}. 
%This respondent had experience in only one shared task. *** CHECK THIS

\subsubsection{Organiser participation in shared tasks (PI-4)}
\label{sssec:organiserParticipation}

%\texttt{"As a participant, it was made clear to me whether or not organisers were allowed to participate in the shared task"}\\

\begin{figure}[h!]
\begin{center}
\includegraphics[scale=0.65]{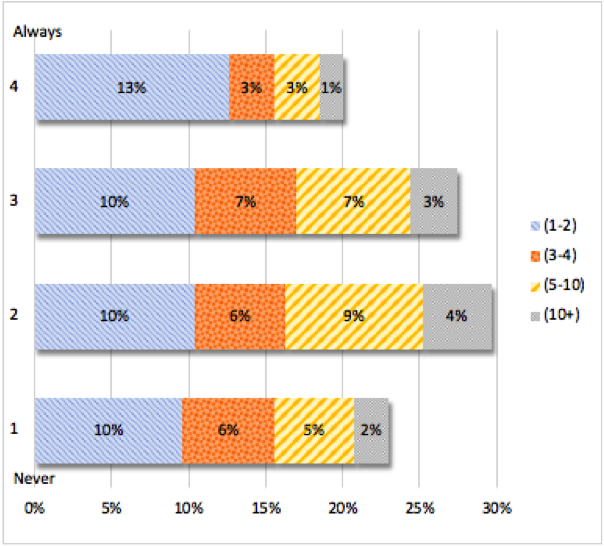}
\caption{\texttt{"As a participant, it was made clear to me whether or not organisers were allowed to participate in the shared task"}.}
\label{fig:transparencyOrgsParticipate}
\end{center}
\end{figure}

As some shared tasks allow for organisers' participation, we asked whether, in the respondents' experiences, this information had been made clear. 
%illustrates how respondents perceived whether or not it was clear if the organisers of a shared task would be participating in said shared task. 
The answers to this question were provided on a Never -- Always scale of 4 (see Figure~\ref{fig:transparencyOrgsParticipate}). The responses in this instance vary widely, with no clear consensus as to whether it is common practice in shared task organisation to clarify this detail for participants. Many of those who provided comments alongside a \textit{Never} selection were not in fact sure, but had \textit{assumed} this was the case. Interestingly, many of the comments from those with middle-ground responses also echoed the same uncertainty and assumption. \footnote{Note that the number of comments relating to uncertainty suggests that this question would have benefited from a ``Not Sure" option.}

\subsubsection{Annotator participation in shared tasks (PI-2, PI-4)}
\label{sssec:annotatorParticipation}

%\texttt{"As a participant, it was made clear to me whether or not the shared task data annotators were also allowed to participate"}\\

\begin{figure}[h!]
\begin{center}
\includegraphics[scale=0.65]{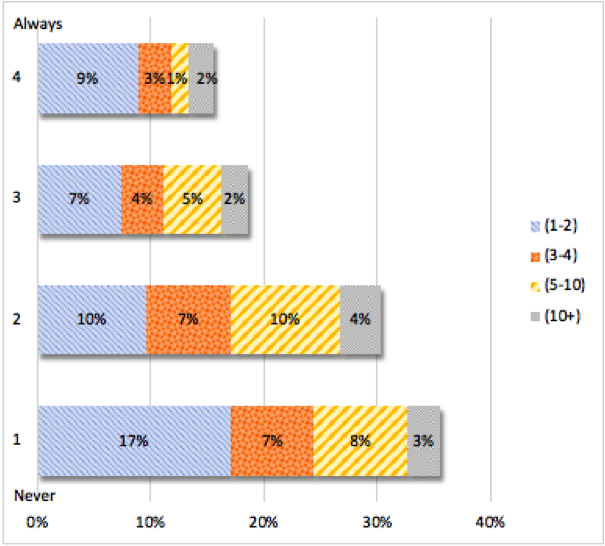}
\caption{\texttt{"As a participant, it was made clear to me whether or not the shared task data annotators were also allowed to participate"}.}
\label{fig:transparencyAnnotatorsParticipate}
\end{center}
\end{figure}

As some shared tasks allow for annotator participation, we asked whether, in the respondents' experiences, this information had been made clear. 
%Next, we enquired our respondents if it was clear to them whether annotators involved in the shared task data collection efforts could participate in the shared task (\ref{fig:transparencyAnnotatorsParticipate}).
The answers to this question were provided on a Never -- Always scale of 4 (see Figure~\ref{fig:transparencyAnnotatorsParticipate}).  While the majority of respondents (66\%) had a tendency towards a negative response, the comments revealed that in this instance, many were uncertain as to whether or not it was in fact made clear at the time. While a ``not sure" comment \textit{suggests} that it had not been made clear enough, we cannot draw this conclusion as there could be a number of reasons for participants not noticing or remembering. \footnote{As with the previous question, it seems that this question would have benefited from a ``Not Sure" response option.}

\subsubsection{Tendency to report positive results (PI-5)}
\label{sssec:posResults}

%\texttt{"As a participant, I have had the tendency to report positive rather than negative results"}

\begin{figure}[h!]
\begin{center}
\includegraphics[scale=0.65]{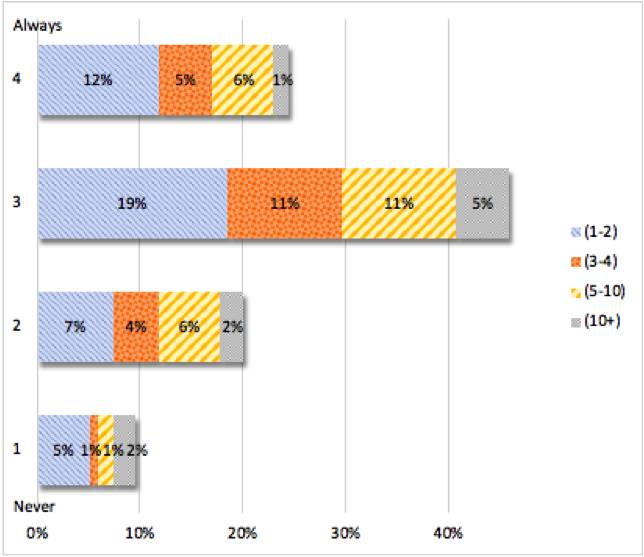}
\caption{\texttt{"As a participant, I have had the tendency to report positive rather than negative results"}.}
\label{fig:ReportPositiveTendency}
\end{center}
\end{figure}

We asked whether, as participants, our respondents tended to report positive rather than negative results. The answers to this question were provided on a Never -- Always scale of 4 (see Figure~\ref{fig:ReportPositiveTendency}). The majority of respondents (70\%) confirmed that their tendency was towards reporting positive results over negative results. A couple of comments referred to the negative impact that reporting negative results can pose when seeking publication. Others, who tend to report both positive and negative results, suggested that shared tasks should be considered the ideal venue to report negative results. One respondent suggested the inclusion of error analysis to make negative result reporting more appealing.

\subsubsection{The link between shared task success and future funding (PI-5, PI-7)}
\label{sssec:futureFunding}

%\texttt{"As a participant, I feel that success in a shared task is important for future funding of my research"}

\begin{figure}[h!]
\begin{center}
\includegraphics[scale=0.65]{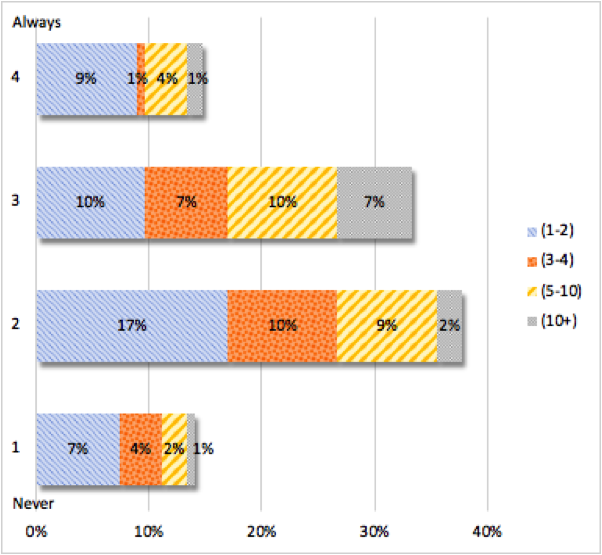}
\caption{\texttt{"As a participant, I feel that success in a shared task is important for future funding of my research"}.}
\label{fig:successAndFunding}
\end{center}
\end{figure}

We also wanted to find out if our respondents felt that success in a shared task is important for future funding. The answers to this question were provided on a Strongly Agree -- Strongly Disagree scale of 4 (see Figure~\ref{fig:successAndFunding}). The results were quite varied, with the minority (29\%) equally divided on strong opinions towards both ends of the scale. The middle-ground responses accompanied by comments suggest that while doing well in a Shared Task can be helpful with respect to gaining future funding, it is not so important or crucial. One ``Strongly Agree" respondent pointed out a concern that when this is the case, there is a risk that participants focus too much on high scores as opposed to exploring new research ideas.

\subsubsection{System tuning towards evaluation metrics vs real-life application (PI-7)}
\label{sssec:metricTuning}

%\texttt{"As a participant, I tend to tune my system towards the evaluation metrics rather than to the real world application of the system"}\\

\begin{figure}[h!]
\begin{center}
\includegraphics[scale=0.65]{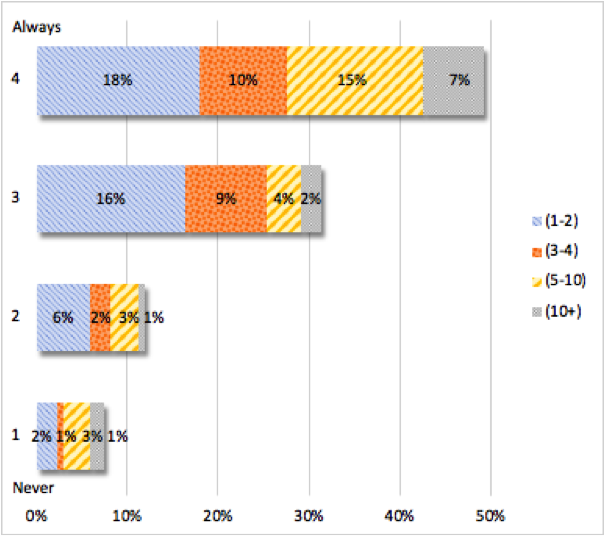}
\caption{\texttt{"As a participant, I tend to tune my system towards the evaluation metrics rather than to the real world application of the system"}}
\label{fig:tuningTowardsMetric}
\end{center}
\end{figure}

 On a Never -- Always scale of 4, we asked respondents if they had a tendency to tune their systems towards the evaluation metrics used in shared tasks, as opposed to real world applications (see Figure~\ref{fig:tuningTowardsMetric}). A clear majority (66\%) confirm a tendency towards this practice. In fact, 40\% report to always doing so.

 Most comments received for this question (13) were included to defend the \textit{Always} response. Many of these stated that the motivation for participating in shared tasks was to do well, or to win, and therefore tuning to evaluation metrics was necessary. Some also pointed out that for some tasks it is difficult to define a real-life application.

\subsubsection{Unequal playing field (PI-8)} 
\label{sssec:unfairAdvantage}

%\texttt{"I think that well-resourced research groups have an unfair advantage in shared tasks"}

\begin{figure}[h!]
\begin{center}
\includegraphics[scale=0.65]{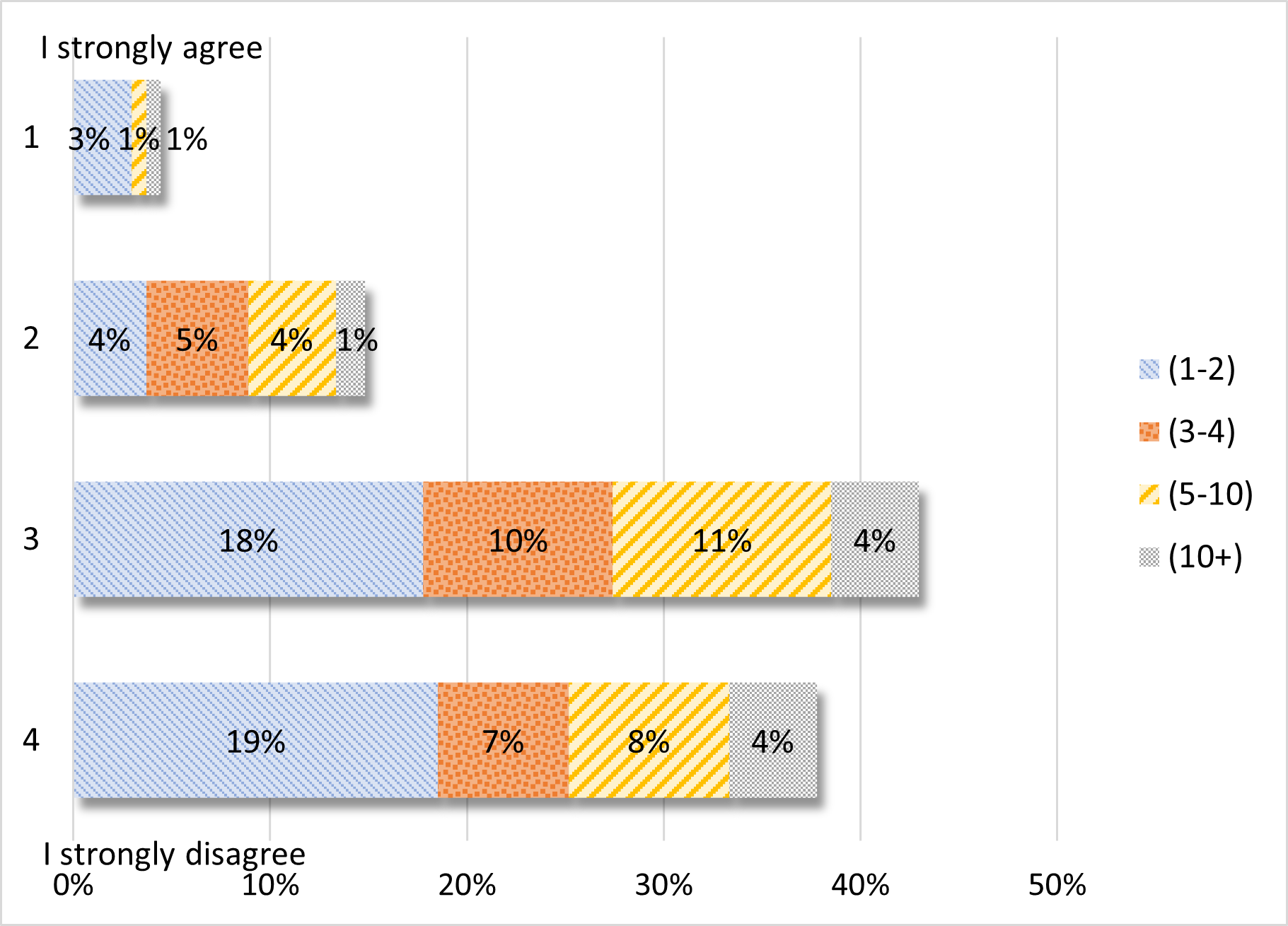}
\caption{\texttt{"I think that well-resourced research groups have an unfair advantage in shared tasks"}.}
\label{fig:unfairAdvantage}
\end{center}
\end{figure}

We also sought to ascertain whether respondents felt that well-resourced research groups have an unfair advantage and we measured this on a Strongly Agree -- Strongly Disagree scale of 4 (see Figure~\ref{fig:unfairAdvantage}).
This question was designed to ascertain whether or not respondents believed that there was an unequal playing field by allowing both modest academic groups and well-resourced researchers (e.g. industry-based) to participate in the same Shared Task.

In the survey, we clarified that by `resource' we meant in terms of both human resources and computing power capabilities.
A strong majority (81\%) agreed with this statement, with almost half of those in strong agreement. Most of the comments provided in this section were provided by those who agreed with the statement (18 comments). These comments show that the question prompted varying opinions on the factors involved leading to an unequal playing field: computational power, human power, time restrictions and data access.  Comments to support those in disagreement varied. Some pointed out that it can be useful to compare research lab systems with commercial ones. Two respondents who chose \textit{Disagree} stated that they did in fact agree with the statement but believed that shared tasks should not be thought of as competitions, and thus ``unfairness" should not apply, while others noted that their response depended on the task and that in some cases they would in fact agree with the statement. 

\subsubsection{Information sharing (PI-1, PI-3)} 
\label{sssec:infoSharing}

%\texttt{"In my experience, participating teams share sufficient information about their systems in an open and timely fashion."} 

\begin{figure}[h!]
\begin{center}
\includegraphics[scale=0.65]{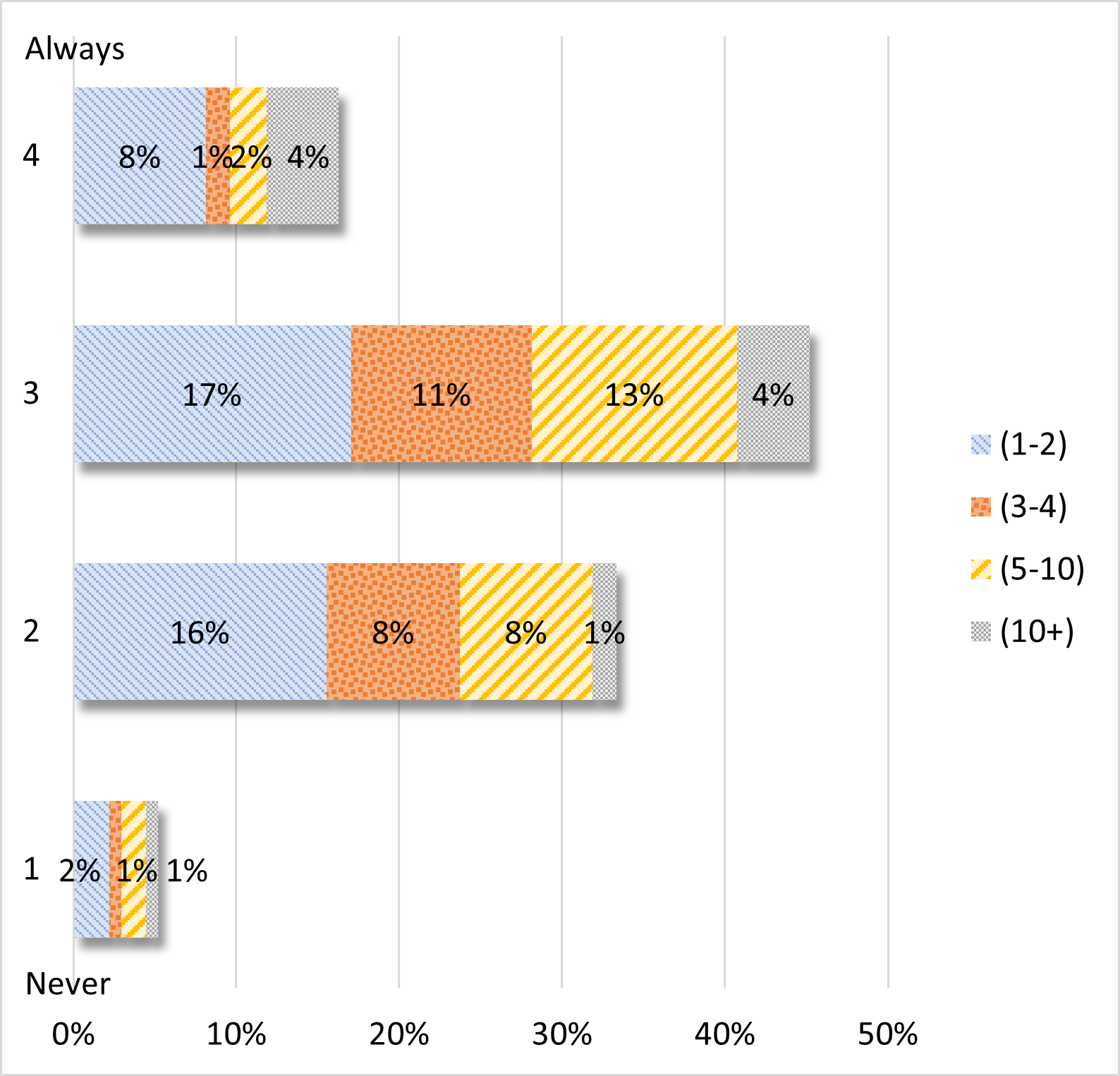}
\caption{\texttt{"In my experience, participating teams share sufficient information about their systems in an open and timely fashion."}.}
\label{fig:infoSharingPax}
\end{center}
\end{figure}

We set out to gauge the respondents' opinion on the manner on which teams reported on their systems. The answers to this question were provided on a Never -- Always scale of 4 (see Figure~\ref{fig:infoSharingPax}). There was little tendency towards strong opinions either way, with the majority (79\%) providing middle-ground opinions. 

Overall, regardless of absolute response, the comments shared a common theme. Many remarked that the answer depended on the shared task. Some highlighted the fact that industry or government participants tended to be more limited or opaque in their reporting, possibly due to copyright or data protection issues. Two respondents noted that the page limit for system description papers (e.g. 4 pages for some shared tasks) was too tight to allow sufficient reporting of the work -- resulting in the need for further documentation through a README file with the code release. Across the board, many of the comments referred to the insufficiency of system information for purposes of reproducibility.

\subsubsection{Replicability in shared tasks (PI-3)} 
\label{sssec:replicability}

%\texttt{"I perceive a lack of replicability of results in some shared tasks."} 

\begin{figure}[h!]
\begin{center}
\includegraphics[scale=0.65]{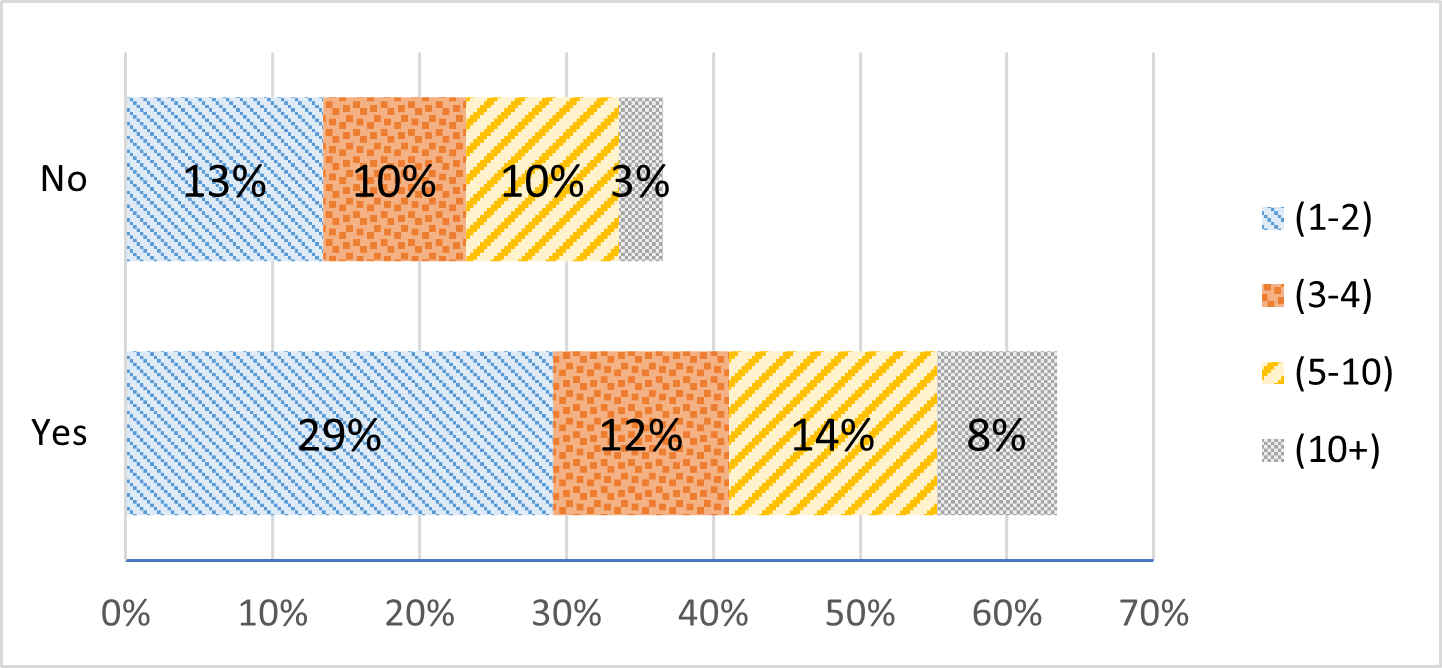}
\caption{\texttt{"I perceive a lack of replicability of results in some shared tasks."}}
\label{fig:replicabilityPax}
\end{center}
\end{figure}

This Yes/No question was aimed at assessing the broad opinion of participants on the level of replicability of results that they observed (see Figure~\ref{fig:replicabilityPax})%\footnote{Note that from this point on, we capture different metadata in presenting our results. For these questions, we have deemed it interesting to have an insight into whether organisational experience impacts responses.
%we use in our analysis a different combined approach of showing responses to a particular question. As we deemed it interesting to see if experience in organizing a shared task has an impact, we now weigh in the respondents' experience participating and/or organizing shared-tasks.}.

It is tied closely to the previous question but specifically targets the question of reproducibility. There was a clear (64\%) majority in agreement with this observation. No comment section was provided with this question, but this was partially compensated by the comments provided for the previous and following questions (cf. Subsections \ref{sssec:infoSharing} and \ref{sssec:replicability}).

\subsubsection{Issues relating to replicability (PI-3)} 
\label{sssec:replicabilityIssues}

%\texttt{"I believe that a lack of replicability is problematic."} 

\begin{figure}[h!]
\begin{center}
\includegraphics[scale=0.65]{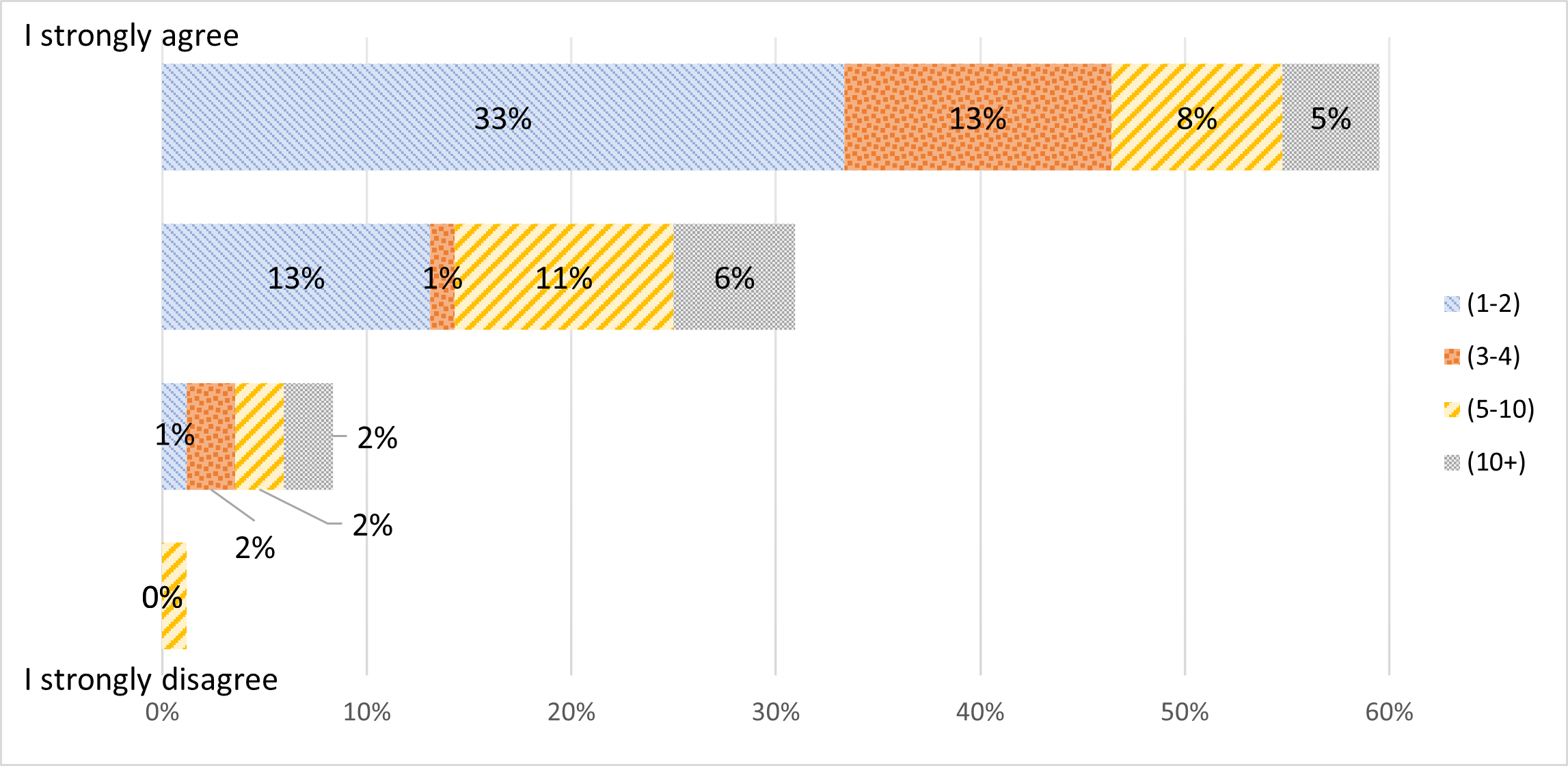}
\caption{\texttt{"I believe that a lack of replicability is problematic."}.}
\label{fig:replicabilityPax2}
\end{center}
\end{figure}

As a final question regarding the participation in shared tasks, we asked our respondents about their opinion on the importance of replicability. The responses to this question were captured on a Strongly Agree -- Strongly Disagree scale of 4 (see Figure~\ref{fig:replicabilityPax2}). Note that due to a glitch in the survey, only 89 people responded to this question. The overwhelming majority (81 out of 89 (91\%)) confirmed that they believe that the lack of replicability is problematic.

Those who disagreed pointed out that shared tasks are meant to inspire research, not to build tools, and  also acknowledged in terms of data replicability that even human consensus is hard when annotating especially when it is subjectively driven (e.g. humour).
Those who responded with ``Strongly Agree" believe that science should mean reproducibility and that technical solutions such as Docker and Anaconda could help to improve replicability in the future. Others pointed out helpful measures that are in place by some existing shared tasks such as the growth of datasets incrementally over the years.

%\vspace{0.5cm}
\newpage
\subsection{Questions relating to the organisation of NLP shared tasks}
\label{ssec:organisationResults}

Of the 167 respondents, 65 of them had previously been an organiser of a shared task and 102 respondents had not. As we deemed it important to also understand the perception towards the overall organisation of shared tasks, all respondents were asked to reply to this section too. For these questions, we have deemed it interesting to have an insight into whether organisational experience impacts responses. Thus, the organisation experience breakdown is captured as metadata in the results we present for the answers received to this section of the questionnaire. 

\subsubsection{Participation of organisers in their own shared task (PI-4)}
\label{ssec:organisersOrg}

As a complementary question to the one in Section~\ref{sssec:organiserParticipation}, where we asked respondents about their experience as participants in shared tasks where the organisers participated in their own shared task, we now gauged opinions on organisers participating in their own shared task from an organiser's perspective.  The answers to this question were provided on a Never -- Always scale of 4. However, it should be noted that based on a closer analysis of the comments we obtained for this question, we observed that respondents may have responded in terms of the \textit{extent} to which they ``agreed" or ``disagreed" to the statement, that is, in contrast to the Never -- Always scale provided e.g. ``I am strongly against this point as this may lead to a biased result").\footnote{While this is a question that should perhaps have been worded differently, the accompanying comments suggest that we have captured the correct sentiment.}

\begin{figure}[h!]
\begin{center}
\includegraphics[scale=0.65]{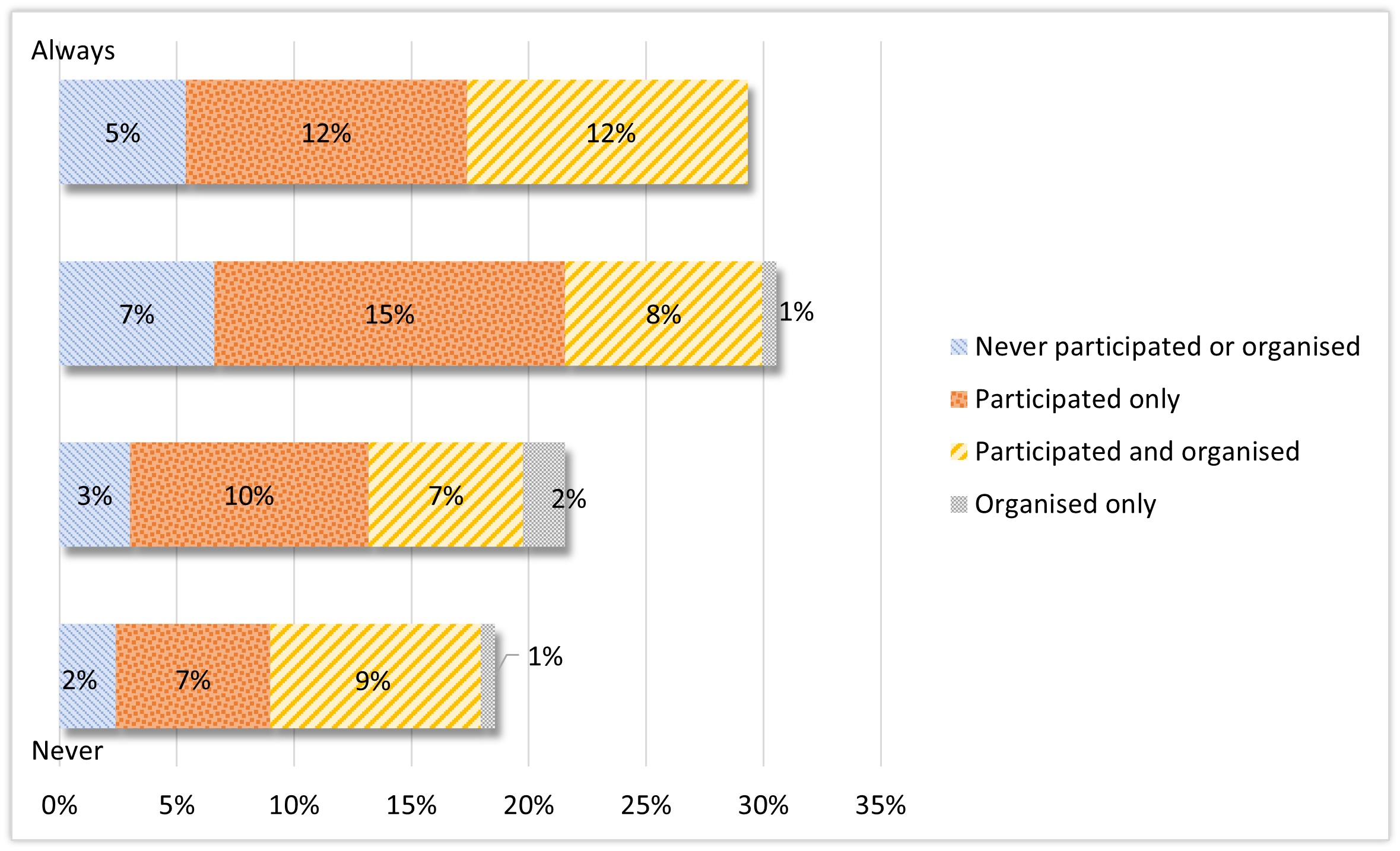}
\caption{\texttt{"I think shared task organisers should be allowed to submit their own systems"}}
\label{fig:organiserParticipationOpinion}
\end{center}
\end{figure}

Figure~\ref{fig:organiserParticipationOpinion} shows how the responses indicate split views on the matter, with an inclination towards \textit{Always} by 60\% of the respondents, against the 40\% that leaned towards the \textit{Never} end of the Likert scale. If we sum up the respondents in the middleground, it shows that about half of the respondents (53\%) do not have a strong opinion on this matter.

The fact that 43 comments were provided with responses to this question shows the need felt by the respondents to back up their chosen point in the scale with further insights. In general, some of those \emph{against} the participation of shared task organisers made points that a baseline score should be provided, but that any other systems submitted by the organising committee should not be ranked. Too much knowledge of the data was cited as the reason for this. On the flip side, there were also comments suggesting that not allowing shared task organisers to participate could discourage involvement in their organisation, and that if proper procedures were put in place to ensure their transparency, then it should be allowed. In relation to this, there were comments suggesting that the terms of participation should be duly reported in the findings paper that typically summarises the results of a shared task, as well as on the shared task website. In total, 12 of the 43 comments received (28\% of all comments) mentioned the need for transparency. In most of those comments, respondents additionally link the role of trust in shared task participation to the need for transparency.

\subsubsection{Participation of annotators in shared tasks (PI-4)}
\label{ssec:annotatorPartOrg}

\begin{figure}[h!]
\begin{center}
\includegraphics[scale=0.60]{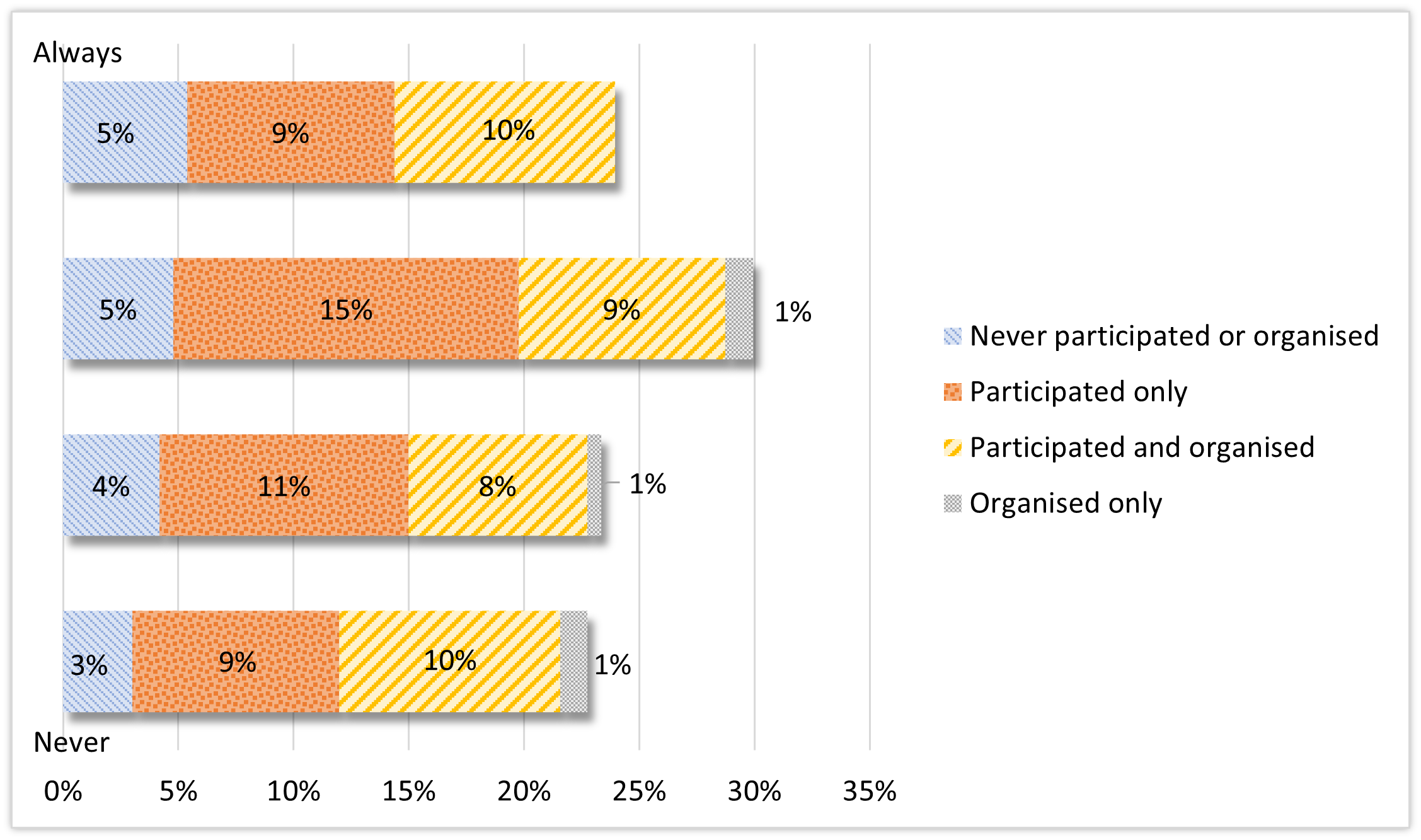}
\caption{\texttt{"In the case of shared tasks that involve annotators, I think that the annotators should also be allowed to submit their own systems"}}
\label{figure:annotatorParticipationOpinion}
\end{center}
\end{figure}

%As certain shared tasks require annotated data and hence organisers engage annotators, we asked our respondents their view on this matter. 
Relating to the question of annotators' participation in shared tasks (c.f. Section~\ref{sssec:annotatorParticipation}), we asked respondents for their view on this. Respondents answered on a Never -- Always scale of 4. As shown in Figure~\ref{figure:annotatorParticipationOpinion}, we observe that the majority of the respondents tend to be in the middle-ground, but leaning towards \textit{Always}. %In fact, 30\% of the respondents chose 3 out of 4 points in the scale as their preferred answer. 
The 36 comments received for this question explained the tendency to not have a clear-cut answer, as many suggested that this would depend on the specific shared task. Again, issues related to transparency and disclosure were predominant.

There were a couple of suggestions to make the annotation guidelines publicly available. In the case of those who selected \textit{Always}, the comments revealed that those respondents did not fully understand the potential underlying issues such as unconscious bias (e.g. ``why not ? in a multilingual system, it's the best way to compare generic system to highly optimised system tailored for one language"). As no follow-up interviews have been carried out (the survey was fully anonymous), it is impossible to determine whether these respondents had not fully considered the implications of annotators participating in shared tasks or if it merely was of no concern to them.%\footnote{From here onwards we no longer distinguish whether the respondents had any shared task organizing experience and focus solely on the responses received.}

\subsubsection{Public release of shared task data (PI-1, PI-3)}
\label{ssec:releaseOfData}

\begin{figure}[ht!]
\begin{center}
\includegraphics[scale=0.60]{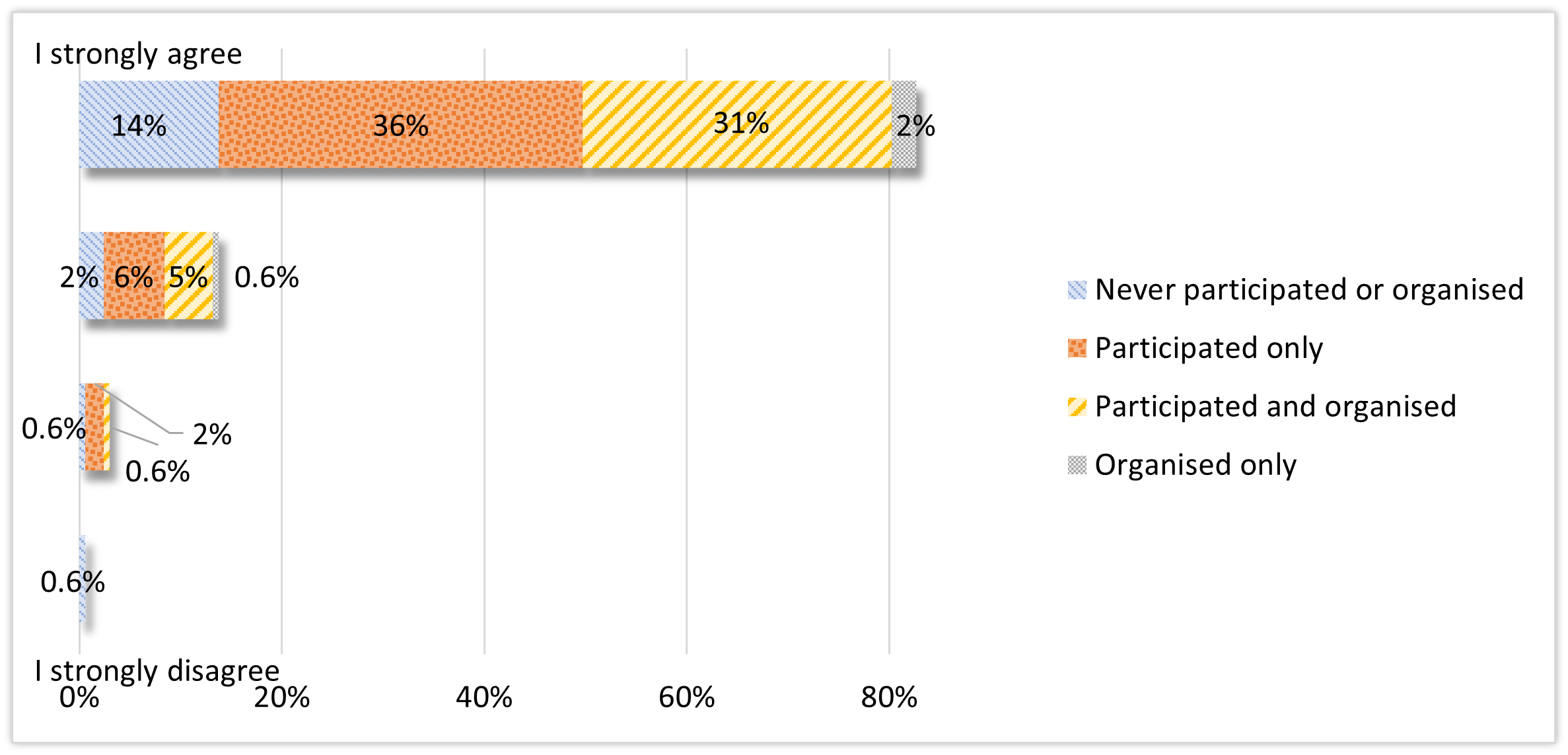}
\caption{\texttt{"I believe that the data of shared tasks should be publicly released upon completion of the shared task itself"}}
\label{fig:releaseOfData}
\end{center}
\end{figure}

To understand respondents' opinion on the importance of sharing data, we asked whether the data should be made available once the task is over. The answers to this question were provided on a Strongly Disagree -- Strongly Agree scale of 4 (See Figure~\ref{fig:releaseOfData}). 

The majority of our respondents (83\%) selected \textit{Strongly Agree} on the Likert scale, with only 1 respondent strongly disagreeing and 5 indicating some sort of disagreement, if we interpret the second point in the scale as \textit{disagree}. Some of the respondents indicated through their comments that while they were generally in agreement with this statement, they were also aware of the fact that measures should be taken for certain datasets, and that in some cases, the release of the data may not be possible at all for reasons of copyright or confidentiality. Insightful hints were offered towards addressing privacy issues if data were to be released (e.g. normalisation and anonymisation). 

Finally, we observed disagreement as to what type of data should be shared. While some of the comments suggested that all data used in a given shared task should be publicly released, others believed that the test set should not be shared.

\subsubsection{Sharing of evaluation metrics (PI-2)}

\begin{figure}[h!]
\begin{center}
\includegraphics[scale=0.55]{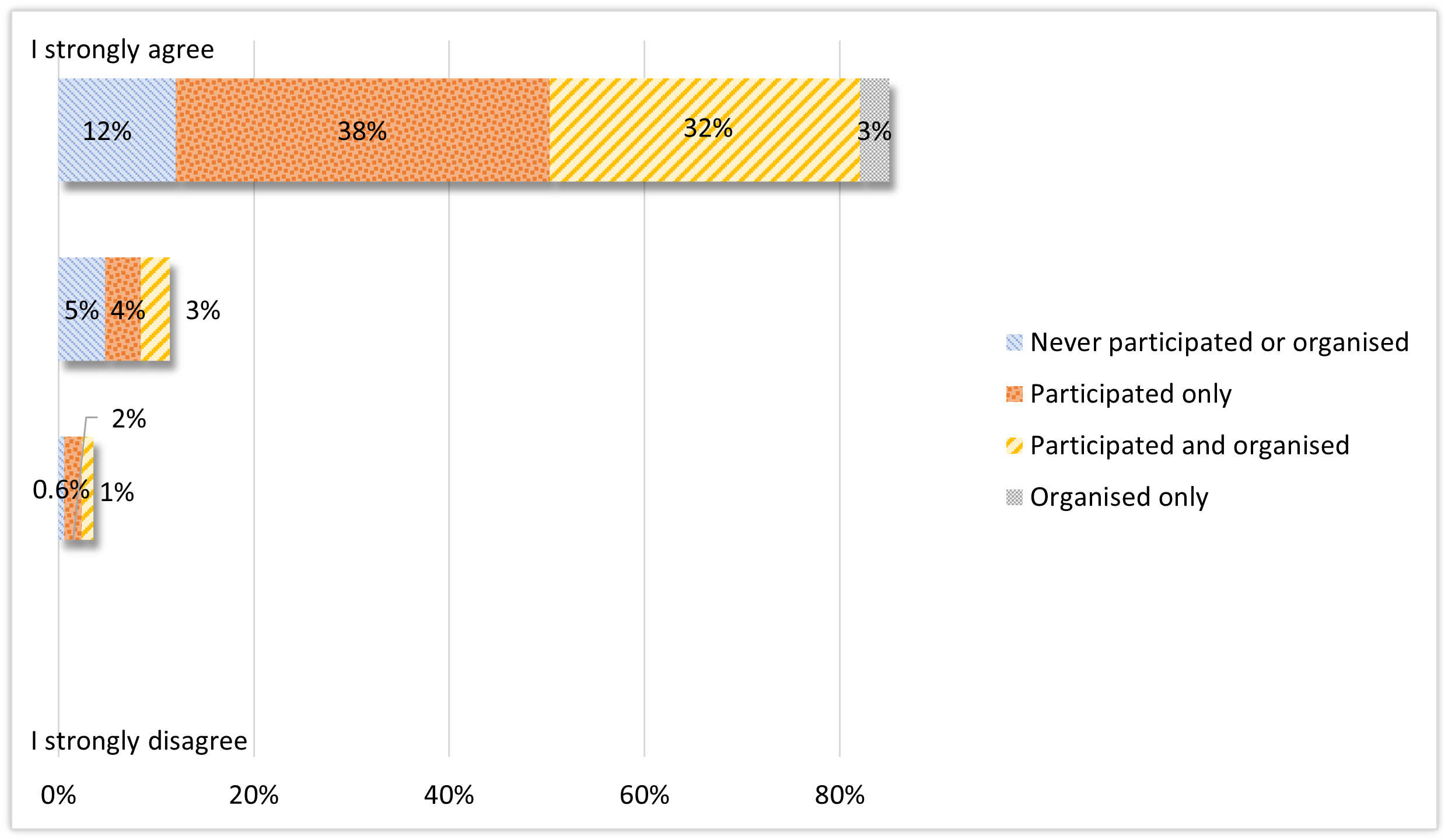}
\caption{\texttt{"I think that the evaluation metrics that will be used to evaluate the shared task should be clearly stated from the announcement of the shared task"}}
\label{fig:metricsInfo}
\end{center}
\end{figure}

We asked whether it was necessary to share evaluation metrics with participants from the outset. The responses to this question were captured on a Strongly Disagree -- Strongly Agree scale of 4. As can be observed in Figure~\ref{fig:metricsInfo}, the vast majority of our respondents (85\%) strongly agreed with this, nobody strongly disagreed, and only 6 respondents showed some sort of disagreement. Although we only received three comments from those who disagreed, two of those indicated that their opinion would depend on the shared task itself, such as in shared tasks where the metrics are actually being defined (such as benchmarking workshops). %If the aim of the shared task is to establish new evaluation metrics, announcing the metrics to be used would not be even possible. 

Among those who agreed, one respondent mentioned that sometimes metrics change over the course of the shared task but that it should be still be possible to share them. There were also comments relating to the fact that releasing the evaluation metrics could sometimes lead to over-fitting. A few respondents argued that the metrics and evaluation procedures should be released to allow participants to perform their own tests and verify the correct data format for submissions. Finally, one respondent highlighted the need for the inclusion of human evaluation with automatic evaluation metrics.

\subsubsection{Stating the type of shared task in the task description (PI-2, PI-8)}

\begin{figure}[b!]
\begin{center}
\includegraphics[scale=0.60]{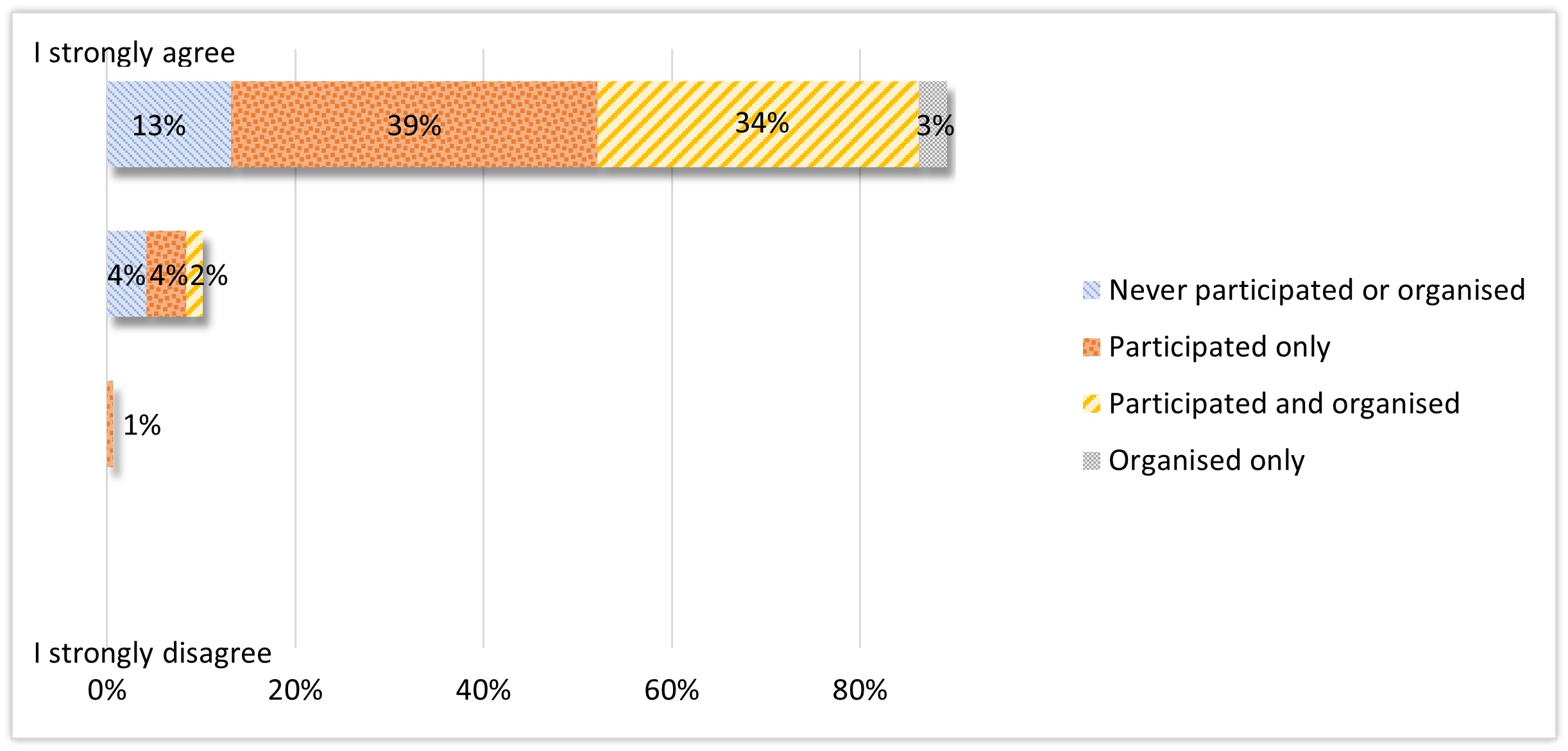}
\caption{\texttt{"I think the type of shared task should always be stated in the task description (open/closed/both)"}}
\label{fig:descriptionType}
\end{center}
\end{figure}

Depending on restrictions defined by organisers regarding data use, the task will be referred to as \textit{open} (the participants may also use their own data) or \textit{closed} (the participants may only use data provided by the organisers). We wanted to ascertain if this distinction of type should be made clear to participants. 

The answers to this question were captured on a Strongly Disagree -- Strongly Agree scale of 4. As shown in Figure~\ref{fig:descriptionType}, almost all respondents agreed, or strongly agreed (89\%) to this statement, with only 1 respondent disagreeing. We only received 8 comments to this question, which may indicate that respondents felt that this is a given in any shared task and there was no need to justify their answers.

\subsubsection{Co-authoring of overview papers of a shared task (PI-2)}

In this question, we sought respondents' views on who should be named as authors on shared task Overview papers. To gauge this, we asked them to select one of the following options:

\begin{enumerate}[label=\alph*)]
\item The shared task organisers only
\item The shared task organisers and the lead annotators (if applicable)
\item The shared task organisers and all annotators (if applicable)
\item The shared task organisers and the participants
\item The shared task organisers, the lead annotators (if applicable), and the participants
\item The shared task organisers, all annotators (if applicable), and the participants
\item Other... (specify)
\end{enumerate}

\begin{figure}[h!]
\begin{center}
\includegraphics[scale=0.65]{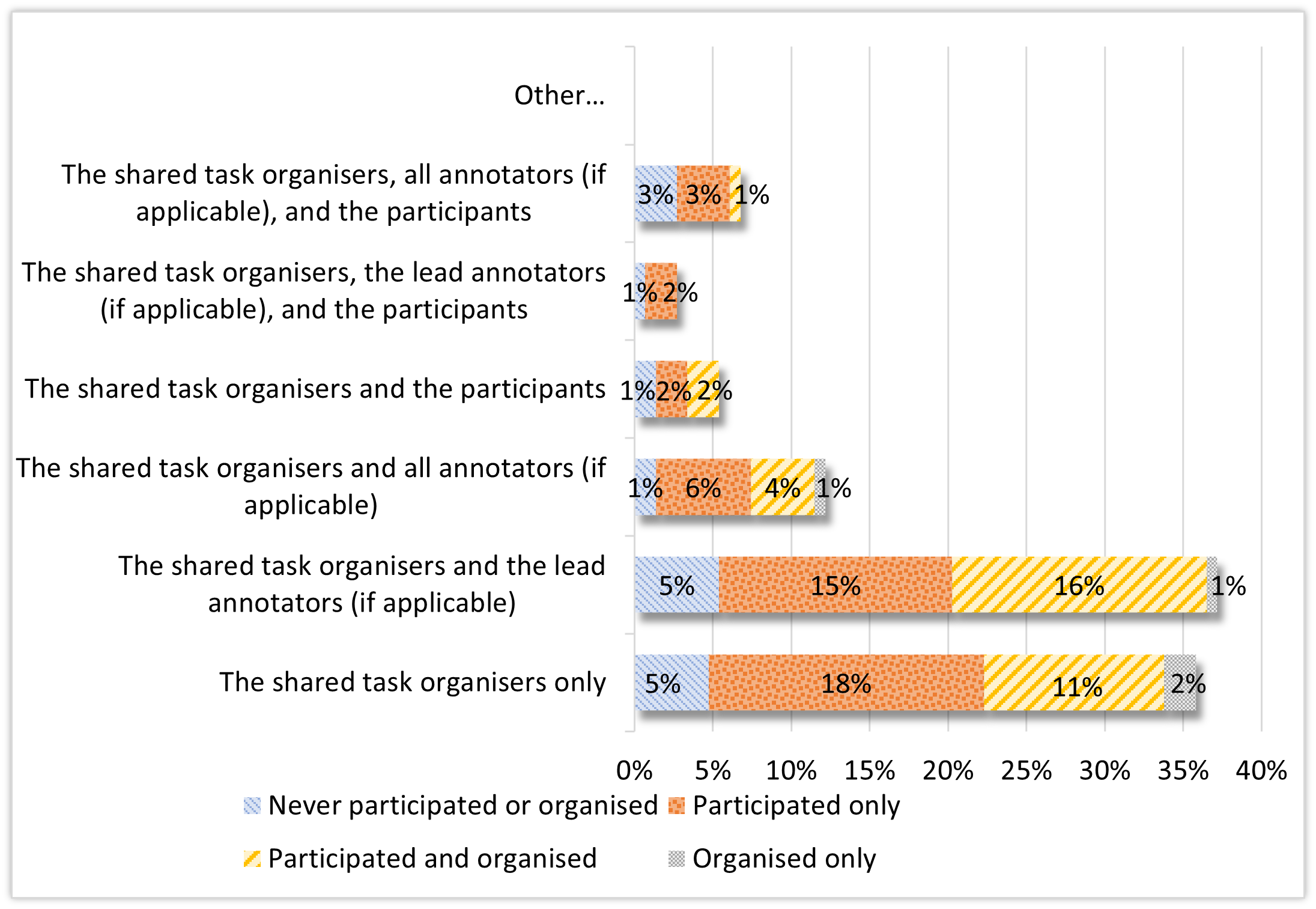}
\caption{\texttt{"Who do you think should be listed as co-author of the OVERVIEW paper of a shared task?"}}
\label{figure:426}
\end{center}
\end{figure}

Unfortunately, due to a bug in the form this question was not marked as obligatory, and only 149 of our 167 respondents replied to it. In any case, and as shown in Figure~\ref{figure:426}, the respondents show a clear tendency towards including all key players in the organisation of the shared task. However, we note a 50/50 divide on whether or not lead annotators should also be included alongside organisers in the overview paper. There was no comment section for this question.

\subsubsection{System descriptions (PI-1, PI-2, PI-3)}

\begin{figure}[h!]
\begin{center}
\includegraphics[scale=0.65]{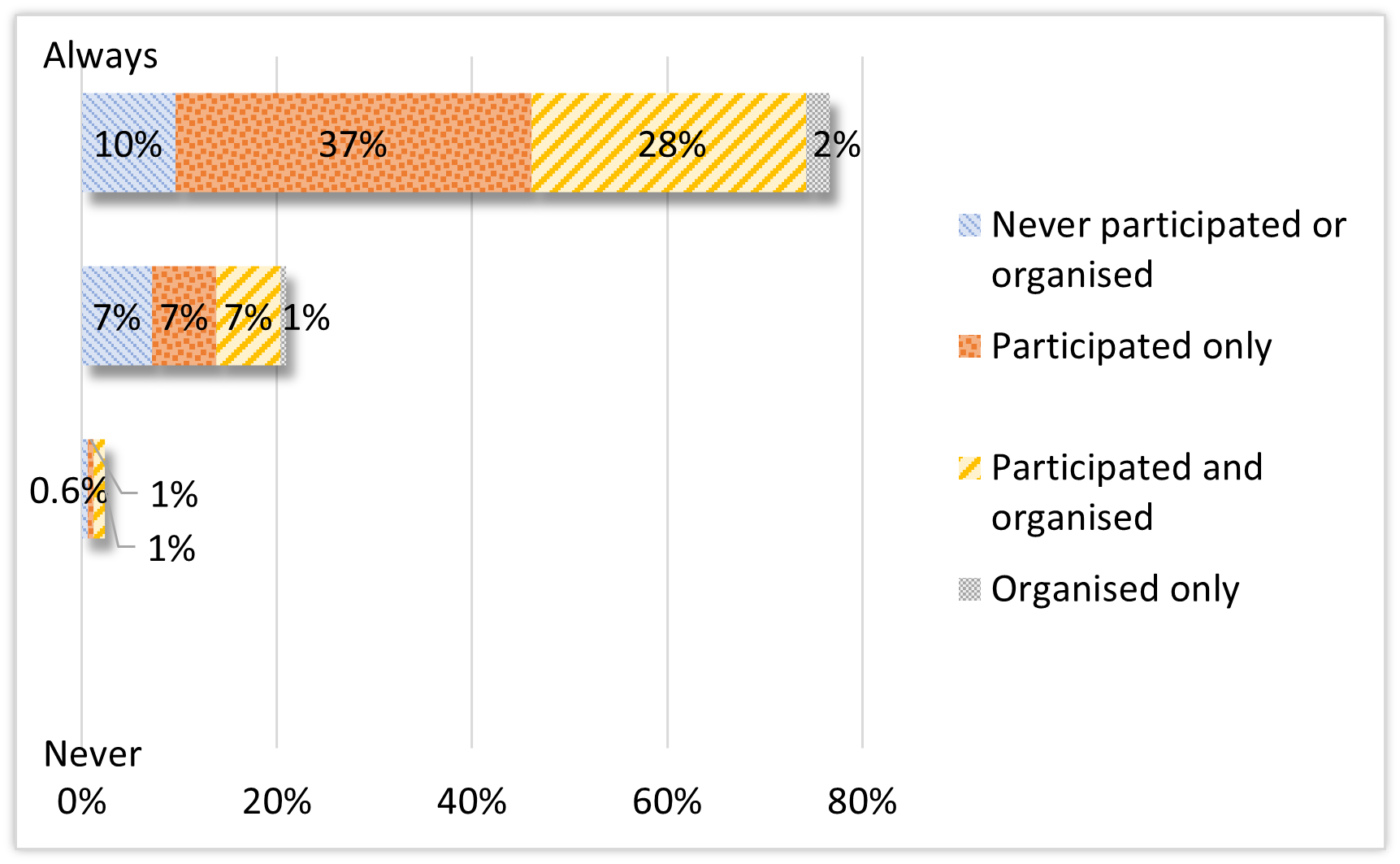}
\caption{\texttt{"In my opinion, participating systems should be duly described in publicly available working notes, papers or articles"}}
\label{fig:SysDescriptions}
\end{center}
\end{figure}

We asked respondents for their opinion on the need to make descriptions of participating systems publicly available. The responses to this question were on a Never -- Always scale of 4. As evidenced in Figure~\ref{fig:SysDescriptions}, there is a clear trend toward an almost unanimous agreement that this should be the case, with only 4 people selecting the option closer to the \textit{Never} end of the scale. Of these 4 respondents, one of them left a comment explaining that in his/her opinion, it would be ``impossible to give a general rule because some systems may be black boxes or commercial".

Among the comments received relating to whether public working notes, papers or articles should be released, many stated that only the best-performing systems need full descriptions, suggesting that those whose systems did not perform well may not be interested or motivated to write a detailed report. Some comments pointed out that it would be beneficial if a shared task made it a requirement to submit system description papers, and there was also a suggestion for organisers to establish guidelines as to how to report the information, keeping it consistent across systems. Several respondents additionally highlighted the need for system descriptions to allow for reproducibility. 
%Finally, one respondent also suggested that both the system code and any additional data used should be released where possible.
These results support the claim that not enough information is currently shared by participating teams to shared tasks (c.f. \ref{sssec:infoSharing}, where it can be seen in Figure \ref{fig:infoSharingPax} our respondents did not feel enough information is currently shared).

\subsubsection{Reporting on additional data used (PI-1, PI-2, PI-3)}

\begin{figure}[h!]
\begin{center}
\includegraphics[scale=0.65]{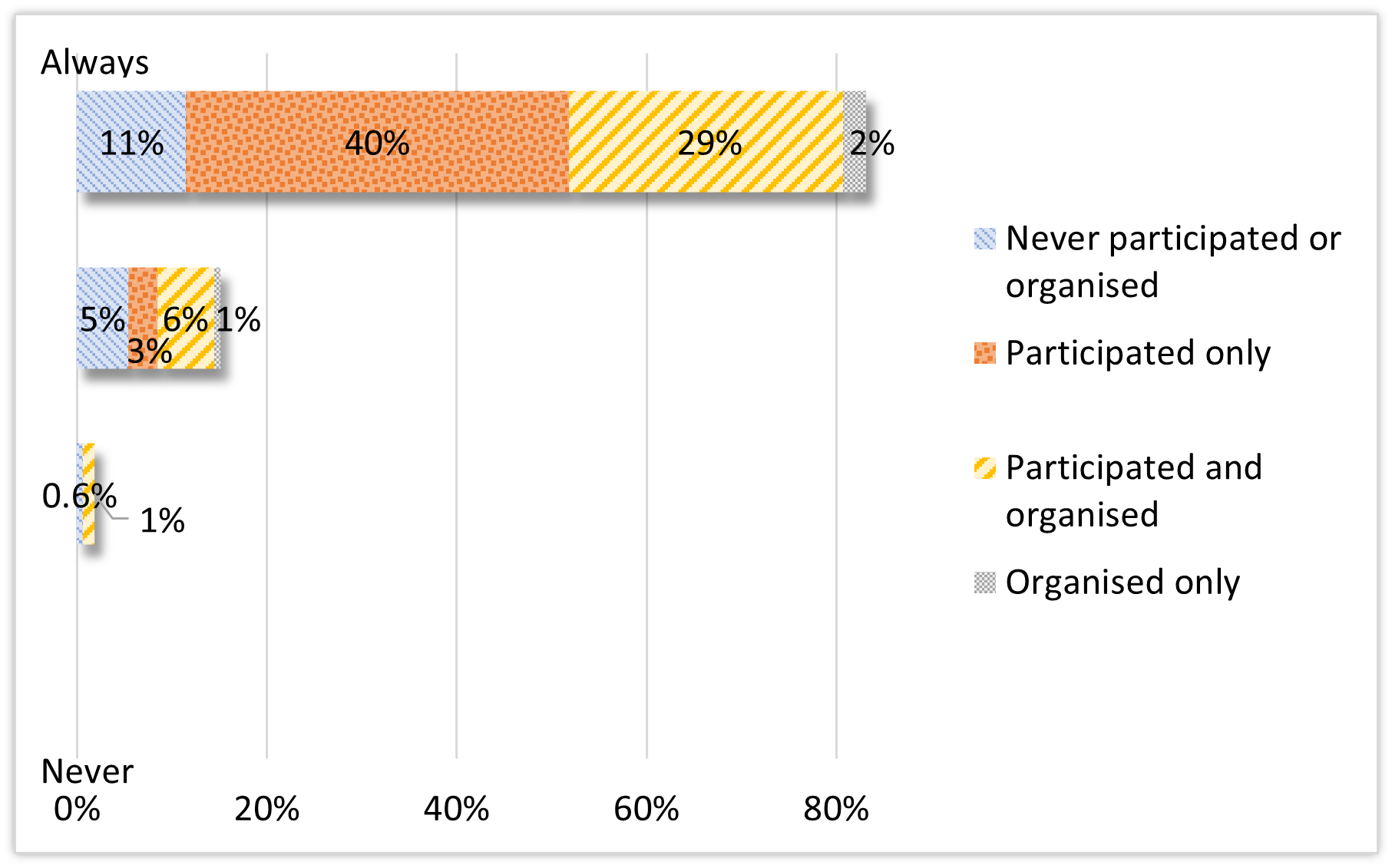}
\caption{\texttt{"In my opinion, teams participating in open shared tasks should report exact details of the additional data they used"}}
\label{fig:reportAddData}
\end{center}
\end{figure}

We also wanted to ascertain the respondents' opinion on the need for reporting on any additional data used by participating teams. Responses to this question were given on a Never -- Always scale of 4.\footnote{Note that for unexplained reasons (possible glitch in the form) while  this question was in fact compulsory, the response of one respondent was lost, hence the total amount of responses being 166 instead of 167.} As was the case with the previous questions, it seems there is a general consensus in the community towards favouring this practice (83\%, c.f. Figure~\ref{fig:reportAddData}), with only three respondents indicating a slight disagreement but leaving no comments to further clarify their reasons for selecting that option.

While this question did not receive many comments (only 17), there was a suggestion to a middle-ground for a ``semi-open'' track where extra data could be used, provided it is open data. This suggestion seems to be in line with other comments regarding restrictions for industry participation (if they were required to release their proprietary data, this could have the caveat of discouraging industry participation). There was also a suggestion for allowing supplementary material in the paper descriptions, as this would make it possible to give better descriptions of the data used.\footnote{It should be noted here that in the last few years, all major conferences in our field allow authors to submit additional materials to supplement their paper submissions. Authors are now encouraged to share data and this is also considered part of evaluation metrics in paper reviews (as part of the reproducibility checks).} Finally, there was another suggestion that organisers provide a template for users to help them describe their data as this could also help content to be more accessible to those browsing through data lists at later stages.

\subsubsection{Involvement of participating teams in human evaluation (PI-4)}

We were also interested in knowing respondents' view on transparency regarding the involvement of shared task participants in human evaluation tasks. For this question, we provided our respondents with three possible answers:

\begin{enumerate}[label=\alph*)]
\item Yes
\item No
\item I have never encountered this situation and therefore I don't have an opinion on it.
\end{enumerate}

\begin{figure}[h!]
\begin{center}
\includegraphics[scale=0.55]{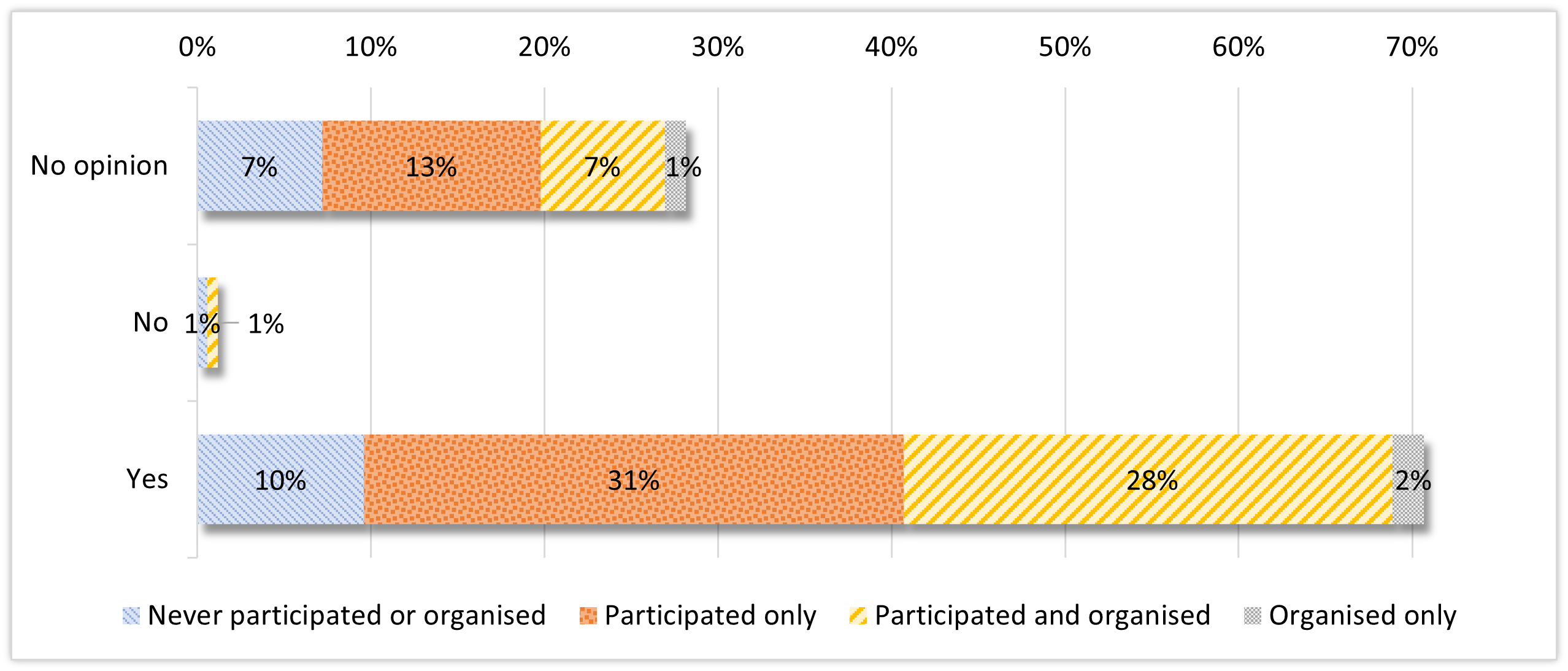}
\caption{\texttt{"I think that it should always be stated up front whether the participating teams will also be involved in human evaluation tasks (e.g. annotating results)"}}
\label{fig:humanEval}
\end{center}
\end{figure}

As shown in Figure~\ref{fig:humanEval}, the majority of our respondents (accounting for 70.6\%) indicated that this should be the case, and 28\% replied that they had never encountered it and hence provided no comments on it.

\subsubsection{Use of a participants' mailing list (PI-1)}
Mailing lists are often a way of facilitating communication between organisers and participants at a group level. Here we wanted to gauge respondents' opinions on the usefulness of this medium.

\begin{figure}[h!]
\begin{center}
\includegraphics[scale=0.60]{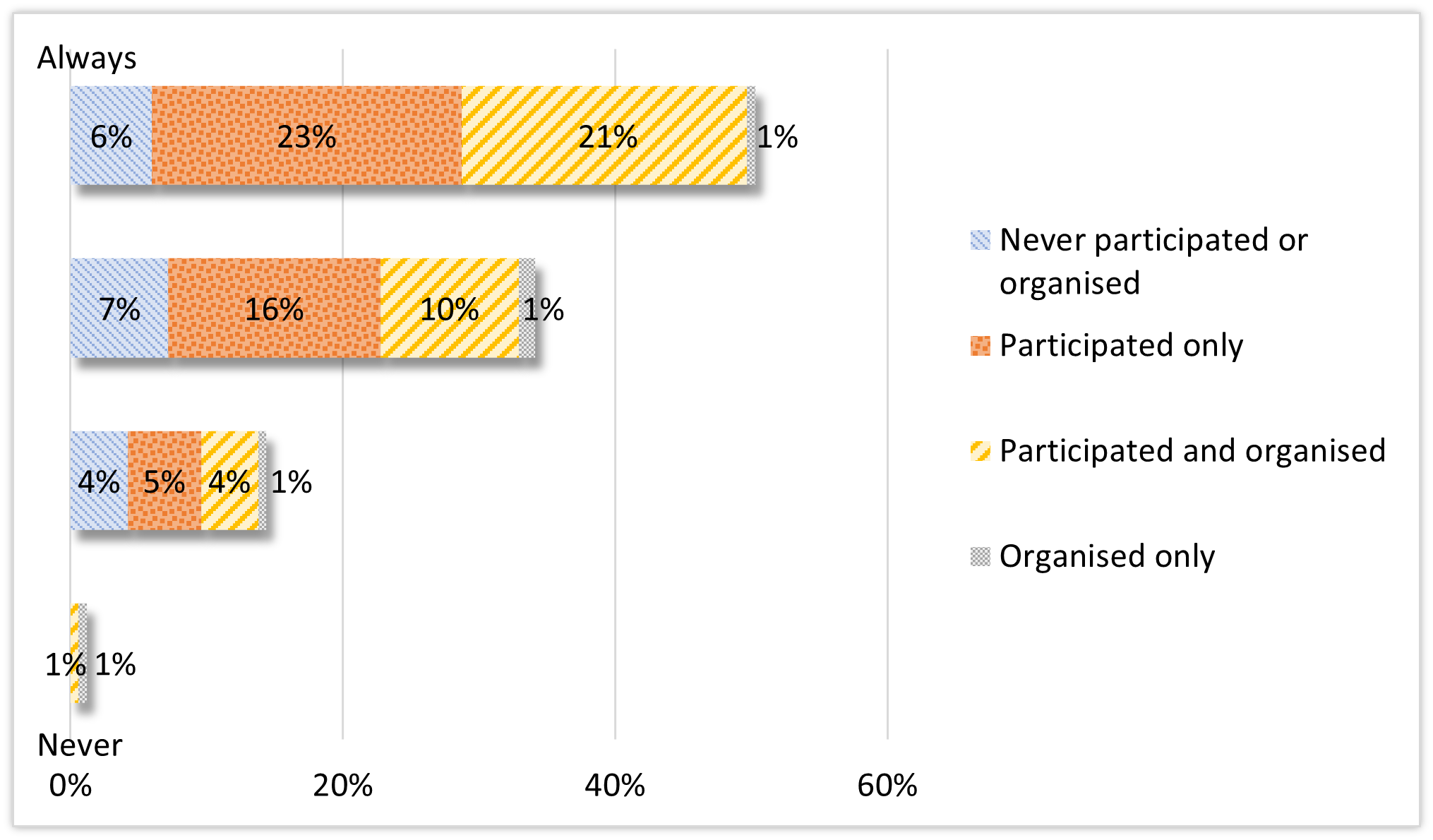}
\caption{\texttt{"Do you think that it is useful to use a participants' mailing list as a common ground for collaboration and sharing of techniques regarding issues when building the systems for a shared task?"}}
\label{fig:mailingList}
\end{center}
\end{figure}

The answers to this question were given on a Never -- Always scale of 4 (see Figure~\ref{fig:mailingList}). Although the answers seem to indicate that 50.3\% of our respondents consider this should be always the case and 34.1\% chose the option closest to always, the remaining 15.6\% leaned towards the \textit{Never} end of the scale. The 23 comments received revealed that this question might have been problematic, as it is not always obvious that the respondents really interpreted what was being asked. Most of the comments accompanying responses that leaned towards \textit{Never} focused on the risk of sharing tips and techniques in the context of a competition environment.

\subsubsection{Differences in the way shared tasks are organised}

\begin{figure}[!ht]
\begin{center}
\includegraphics[scale=0.60]{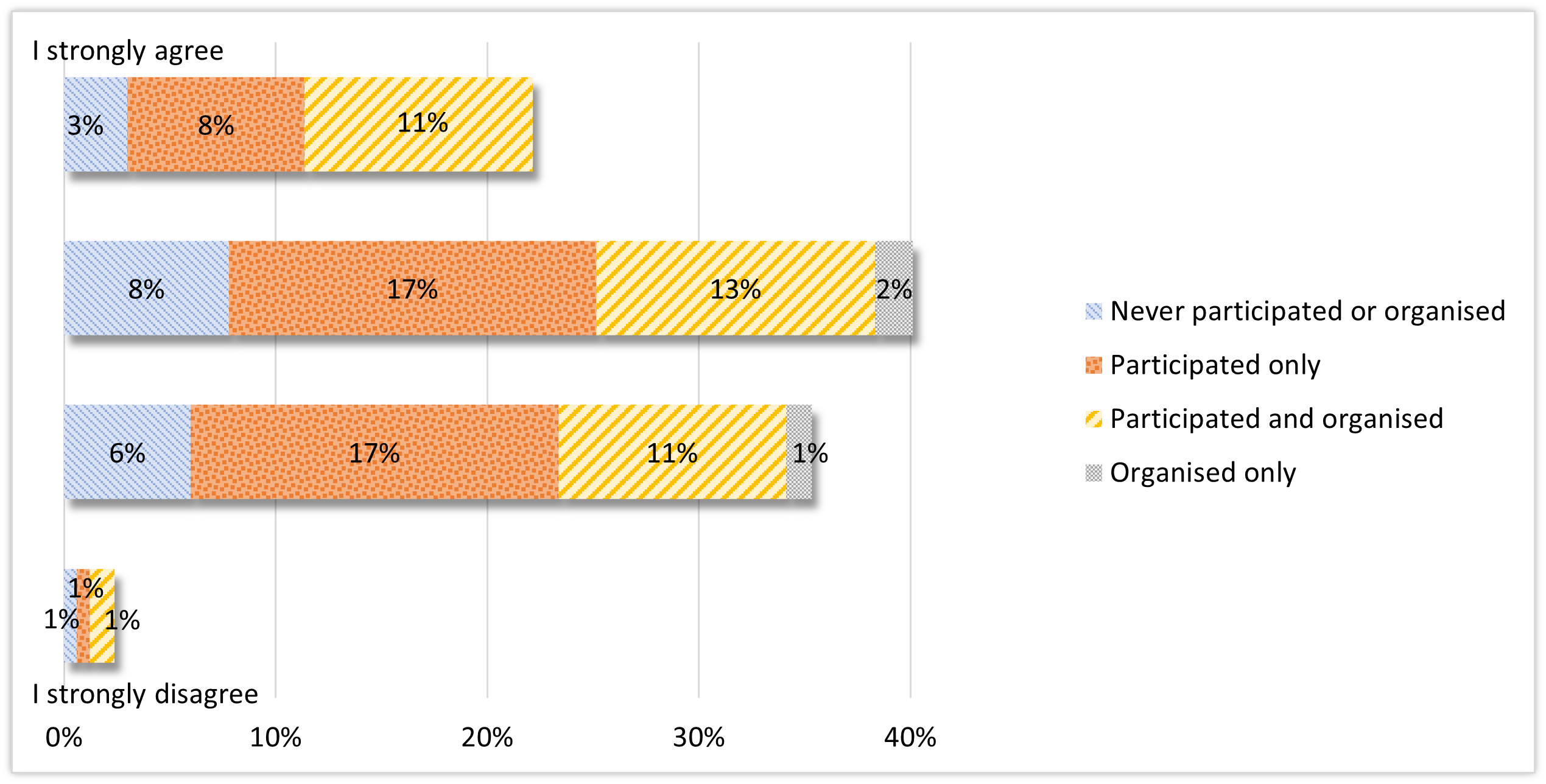}
\caption{\texttt{"In my opinion, there are major differences in the way shared tasks are organised (who can participate, evaluation metrics, etc.) across NLP"}}
\label{fig:Differences}
\end{center}
\end{figure}

Here we wanted to capture respondents' observations on the major differences across shared tasks in terms of their organisation. The answers to this question were captured on a Strongly Disagree -- Strongly Agree scale of 4. As shown in Figure~\ref{fig:Differences}, there appears to be a tendency towards agreeing with this statement (accounting for 62\% if we sum the replies for the two points in the scale on the \textit{agree} end). 

While there were not many comments provided for this question (23), most of the comments showed that our respondents in some cases were not sure of their own reply. This was the case for 9 of the 12 comments in the \textit{disagree} end of the scale (e.g. ``I have no idea''; ``I have not enough experience to give an objective answer''; ``I don't really have an opinion here''). This lead us to conclude that this question should have included the option of answering ``I don't know enough about this to have an informed opinion'', instead of forcing our respondents to take a stance. 

Finally, comments on the agreeing side included pointers to the nature of shared tasks in NLP and suggested that this may be inevitable, ``given the diversity of tasks and organisers''.

\newpage

\subsubsection{Timeline for system training (PI-2, PI-8)}

\begin{figure}[h!]
\begin{center}
\includegraphics[scale=0.60]{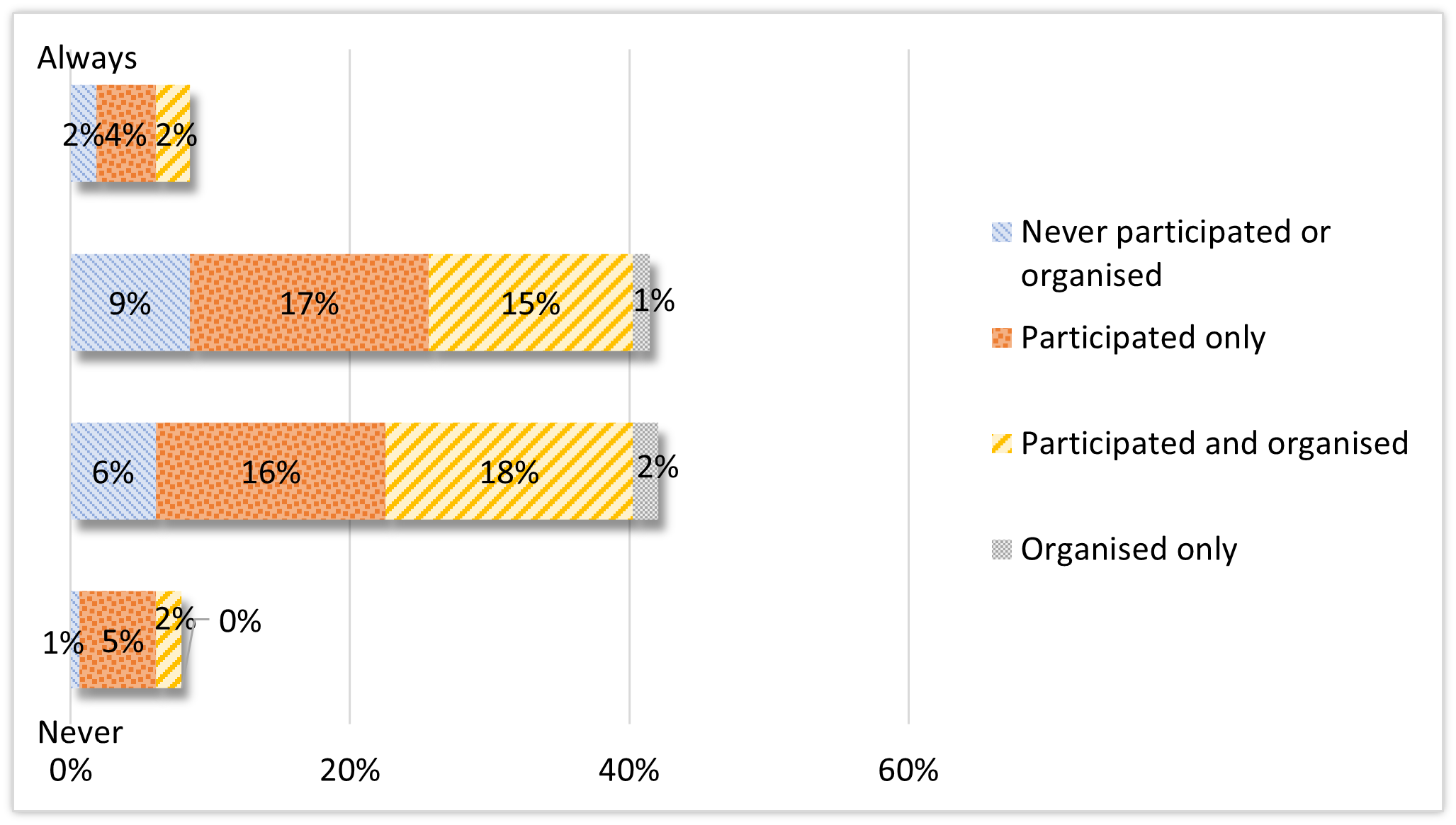}
\caption{\texttt{"In my opinion, the timeline usually given to participating teams to train their systems is too tight"}}
\label{fig:Timeline}
\end{center}
\end{figure}

Timelines provided to participants for training their systems can vary significantly. Therefore we asked respondents if (in their experience) the timelines offered to them were sufficient or too tight. The answer to this question was provided on a Never -- Always scale of 4. This section did not include a comments section, which does not allow us to draw any other insights from the responses obtained other that those indicated in Figure~\ref{fig:Timeline}.

\vspace{-0.5cm}
Additionally, and despite this question being compulsory, it should be noted that the responses of 3 of the respondents were also lost. The responses are mainly distributed among the middle ground (82.4\%), being split into two halves between the \textit{Never} and the \textit{Always} end of the scale.

\subsubsection{Satisfaction with the current organisation of shared tasks}

\begin{figure}[h!]
\begin{center}
\includegraphics[scale=0.60]{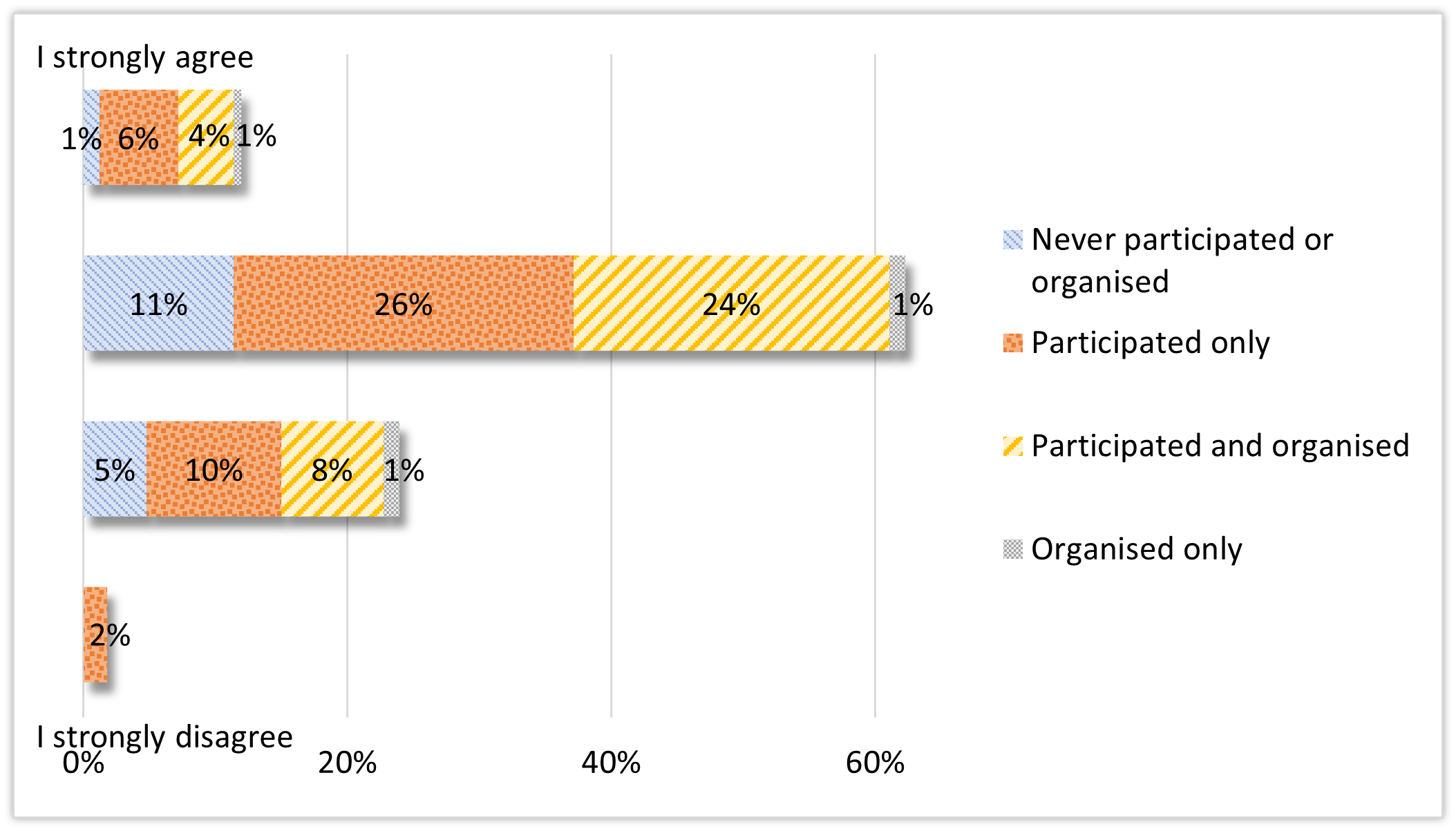}
\caption{\texttt{"I am satisfied with the way shared tasks are organised in NLP"}}
\label{fig:satisfaction}
\end{center}
\end{figure}

Overall, we wanted to gauge respondents' satisfaction with how NLP shared tasks are organised. The answers to this question were captured on a Strongly Disagree -- Strongly Agree scale of 4. Although there seems to be a majority agreeing on this (62.2\%, c.f. Figure~\ref{fig:satisfaction}), upon analysis of the comments received along this question it seems that there is a lot of overlap in opinion between the respondents that selected the option leaning towards disagreeing and the one towards agreeing. In fact, some of those who chose to agree provided comments to highlight that there is room for improvement. Additionally, some of the comments within the agreeing respondents also indicated that they did not know or that their experience was too limited to give a broad objective answer, which lead us to conclude that we should have included an option of ``I don't know'' as a response.

\subsubsection{Common organisational framework}

\begin{figure}[h!]
\begin{center}
\includegraphics[scale=0.60]{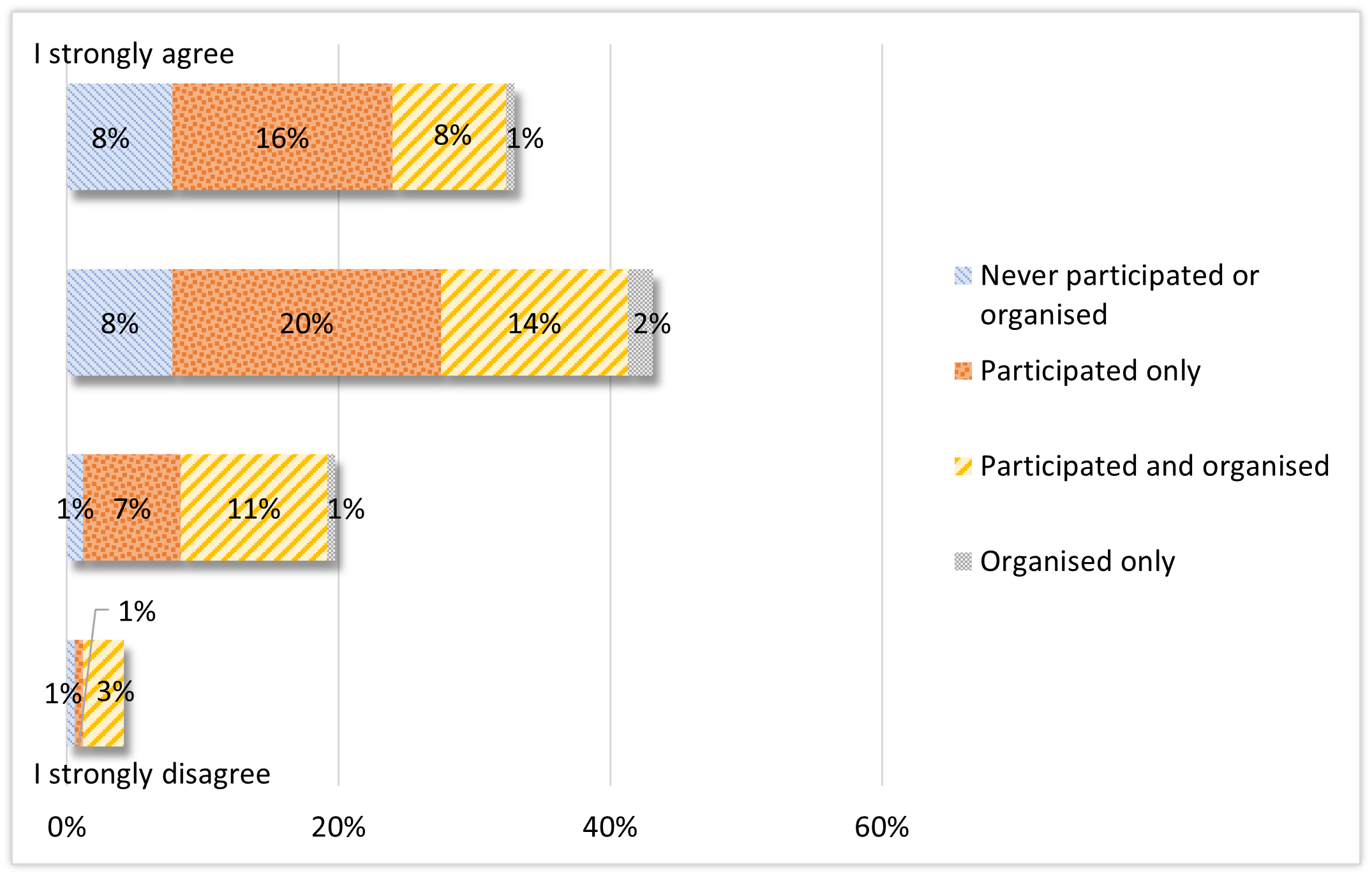}
\caption{\texttt{"I think that shared tasks should share a common organisational framework to help explain clearly to participants what each specific shared task entails"}}
\label{fig:CommonFramework}
\end{center}
\end{figure}

Here we set out to ascertain whether or not respondents saw the need for a common framework which organisers could refer to in the setting up of shared tasks. The answers to this question were captured on a Strongly Disagree -- Strongly Agree scale of 4. As evidenced in Figure~\ref{fig:CommonFramework}, there seems to be a tendency to agree with this statement.

Among the comments received (25), we observed that those agreeing (16 out of 25) with the statement highlighted that it would be good to have a list of important pieces of information that should be provided, and the varying nature of the shared tasks in our field should be accounted for. They also pointed out that sharing best practices would be positive for future organisers and that it would be good to share a common framework or a set of standards for ensuring clarity, consistency and transparency.

Among those disagreeing (9), it seemed that the term ``framework'' was problematic, as respondents had unclear interpretations of it. Some understood it to be a strict set of rules like a ``one-size-fits-all'' set-up rather than a template that can be adapted and used as a guideline. Others noted that it would be hard to visualise how this could be implemented and thus expressed doubts. In fact, other comments disagreeing highlighted the diverse nature of shared tasks as a challenge for implementing it. Finally, some other comments under this category expressed disagreement to sharing a common format for all shared tasks, but supported the idea that common elements in shared tasks could be shared such as best practices. All in all, however, we did not get enough comments to provide insight as to what provoked the disagreement.

\subsection{Final remarks}
\label{ssec:finalremarks}

The last section of our questionnaire consisted of three free-text fields allowing respondents to share any particular (i) positive or (ii) negative experiences encountered with shared tasks, or (iii) any other thoughts we should take into account. These three open-ended questions were aimed at gathering further insights not covered by the questions in our questionnaire and were phrased as follows:

\begin{itemize}
    \item We would really appreciate it if you shared with us any particularly ''positive" experiences you had when participating in and/or organising shared tasks.
    \item We would really appreciate it if you shared with us any particularly ''negative" experiences you had when participating in and/or organising shared tasks.
    \item If you have any further comments, feel free to leave them here :)
\end{itemize}

54 respondents (32.3\%) left comments related to positive experiences, 49 (29.2\%) left comments related to negative experiences, and 21 (12.5\%) left further additional comments. Below, we summarise the positive and negative experiences reported by our respondents.

\subsubsection{Particularly positive experiences}
We received 54 comments in this section and they referred to two main types of experiences related to shared tasks. Some respondents focused on general positive experiences from participating in shared tasks, while others focused on particular shared tasks and their experiences participating in or organising them.

Among the general comments there was a clear tendency towards praising the benefit and sense of belonging to a community or research field that arises from participating in shared tasks. Most respondents also mentioned that shared task participation encourages sharing of ideas, techniques, and resources in a specific area, which motivates participants to carry out further research on that topic. There were also several comments regarding the steep learning curve required to participate in a shared task, but rather than being a negative, this was perceived as a positive opportunity for participants to immerse themselves in a particular field of research, understand things better, build state-of-the-art NLP systems, and move the field forward. Thus, these respondents seemed to perceive shared tasks as a personal challenge for self-learning and development, highlighting the need to think ``how I can apply my hammers to this problem'', or ``how I can improve this task most''. In some cases, respondents also referred to the participation in a shared task as a good team-building exercise, and several respondents mentioned sharing of data and code as positive elements that also foster the sense of community. Finally, one respondent mentioned that shared tasks are positive experiences because researchers get to know not only which techniques work best for a specific task, but also which ones fail, also giving an opportunity to teams not performing well to publish negative results which otherwise are hard to get published.

With regards to positive opinions on how shared tasks were organised, respondents focusing on this praised, in particular, the cases in which organisers were friendly and collaborative, highlighting responsiveness as a key element. There were also references made to positive experiences with shared tasks that had very clear descriptions, high quality data, and stuck to the original timeline. Three respondents additionally referred to evaluation procedures they particularly liked, such as providing the participants with an extrinsic evaluation dataset (CLIN27 shared task), and having both hidden and remote virtual machines where participants had to upload and run their code (e.g. Shallow Discourse Parsing shared task). Some of the other shared tasks mentioned as particularly well organised were the CoNLL 2017 shared task (mentioned twice), the SemEval STS task (mentioned twice), the Sigmorphon 2016, the SemEval 2017 task 2, and the SANCL shared task.

\subsubsection{Particularly negative experiences}
We received 49 comments in this field (although 4 of them were statements to say they did not have any negative experiences). Nonetheless the remaining 45 comments are likely to be informative  to the community with respect to areas that could be developed upon in the future in certain shared tasks. The majority of these responses were from a participant perspective.

In terms of clarity with respect to the shared task description, some respondents reported frustration of the aims and scope of the task changing between the first call and the final call for participation. Some respondents complained about the issue of fairness and cases of participants taking the competition too seriously, instead of seeing shared tasks as a way to advance the field of study.

Complaints were made with respect to teams performing particularly well who neglected to publish their code and were unwilling to provide a system description to help other participants understand what worked best for them. 

 Many respondents highlighted the perceived negative attitude towards publishing negative results as being problematic. They noted that although the competitive nature of shared tasks is intended to move the field forward, it can also have the opposite effect of discouraging the publication of full system descriptions, thereby making work replicable. One respondent summed up the sentiment of others with ``I believe NLP research should be about novel ideas and diligent, replicable work as much as it should be about beating baselines".
%The issue of a negative impact relating to publishing negative results or low ranking was highlighted by many with one respondent summing up the sentiment of others with "I believe NLP research should be about novel ideas and diligent, replicable work as much as it should be about beating baselines".

In terms of poor communication, a number of respondents complained about the issue of organisers changing the evaluation metric in the middle of a shared task.
Similarly a number of complaints were made about the unresponsiveness of organisers to queries from participants. 

%- Demotivation if your system is ranked among the worst performing ones
The issue of recognition in a shared task's Overview paper was highlighted by a respondent who noted that they have never received credit for their annotations on such papers.

Multiple complaints were made in relation to poor data management, such as late release of data, changing the data annotation during training time, and in one case the carelessness of creating test data that differed from the training data was highlighted. Other data-related issues were highlighted such as shared tasks providing many and huge datasets meaning that teams without sufficient computing power could not compete on an equal footing. One respondent pointed to the issue of limited data being released resulting in different data sets for all participants: ``Limited data shared tasks where some data is not accessible to all participants'', and a few others pointed to the fact that sometimes the data sets are not made available after the shared task, preventing possible attempts to advance in the topic and reproduce results, while benefiting those who retain the data.

With respect to timeline management, the point was raised that sometimes there is not enough time allocated to the different phases (preparation, training, testing etc). Also related to timelines, there was the complaint of lack of coordination or synchronisation across related shared tasks in setting their deadlines.

%- proliferation of shared tasks --> hard to get a wide attention from the community and an impact on the community (too many STs on offer leads to less participation). Suggestion to try to synchronize STs and deadlines. 

A number of respondents highlighted the problem of opaque system descriptions lacking the necessary level of detail and hence not allowing for replicability.
 
Finally, there was some advice from organisers of previous shared tasks. For example, they suggest that participants ensure that licensing agreements can allow for earlier sharing of shared task data. In addition, issues arose in the past, where participants did not read instructions carefully resulting in misunderstandings and unnecessary queries to the organisers. On the other hand, some respondents complained about the lack of clarity in the shared task descriptions. This point both raises awareness for the responsibility of the participant in requesting further information where necessary, and reminds organisers that clarity is of the essence when drawing up guidelines or instructions.

\subsubsection{Additional comments}

Nineteen additional comments were provided. A number of these comments thanked the authors for carrying out this study and some others had specific suggestions for the improvement of shared tasks: providing APIs so that systems can be tested remotely, ensuring code is sustainable where possible, and requiring system descriptions provide details of both human and computing resources used. 

Two participants misunderstood the purpose of the survey as being part of a goal to formalise or institutionalise shared tasks. The majority of additional comments provided were observations of the benefit of shared tasks for the field.

\section{Discussion}
\label{sec:discussion}

%"Findings of the E2E NLG Challenge"
%https://arxiv.org/pdf/1810.01170.pdf

%*** A note about the greater awareness for the ethical considerations surround shared tasks and the clear need for a framework. Echoing the results of the survey and the negative/positive discussion in the earlier section JM: Attempted below!**

Recent years have seen a growing awareness of the importance of ethics within NLP, particularly following \citet{hovy-spruit-2016-social}, with a number of workshops and papers devoted to the topic. The contribution by \citet{parraescartin-EtAl:2017:EthNLP} marked the first time that shared tasks became a focus for ethical discussion, and the responses reported in Section \ref{sec:surveyResults} in this article demonstrate that both organisers and participants would like to see a common framework or checklist developed for use in the organisation of shared tasks. Of course, a common framework cannot be expected to cover every eventuality, and individual events may have specific or exceptional requirements that require consideration beyond the remit of such a framework (see \citet{Johannen2020EthicalCO}, for example). However, the provision of a checklist as a basic tool for ethical consideration will no doubt be of benefit to the greater community.

%Along with the feedback from this survey, 
A number of researchers in the field have already taken note of the ethical considerations surrounding shared tasks, as cited by \citet{parraescartin-EtAl:2017:EthNLP,SharingIsCaring}, and more recently, \citet{Ethayarajh2020}. For example, \citeauthor{Dusek2018} (\citeyear{Dusek2018} and \citeyear{DUSEK2020}) took steps towards encouraging the reporting of negative results  where they allowed researchers to withdraw or anonymise their results if their system performs in the lower 50\% of submissions.
%"Utilizing Neural Networks and Linguistic Metadata for Early Detection of Depression Indications in Text Sequences"
%https://arxiv.org/pdf/1804.07000.pdf
Elsewhere, in a shared task on the early detection of depression using machine learning, \citet{Trotzek2018} examine the use of a previously standard metric, and cite the same source noting that ``there are also ethical considerations to keep in mind for NLP shared tasks and shared tasks in general". 
%The competitive nature of such tasks may lead researchers to be secretive about their systems and methods, ethical concerns may be overlooked, and conflicts of interest may arise if organisers themselves participate in a task.

%"Can adult mental health be predicted by childhood future-self narratives? Insights from the CLPsych 2018 Shared Task"
%http://www.aclweb.org/anthology/W18-0614
In addition, the overview paper from the CLPsych 2018 Shared Task \cite{Radford2018} note also how ethical considerations are extremely important not only in shared tasks for clinical NLP and health research but also shared tasks in general, citing the same work by \citet{parraescartin-EtAl:2017:EthNLP}.
%https://archive-ouverte.unige.ch/unige:97428
In the overview of the 2017 Spoken CALL Shared Task, \citet{Baur2017} note that they ``briefly address the issues raised by Escartín and her colleagues, considering conflicts of interest, anonymity, gaming of the system and the balance between competitiveness and collaboration." %\citet{bernier-colborne-langlais-2020-hardeval}, in their investigation of evaluation methods for Named Entity Recognition, refer to the raised issue of machine learning and NLP models failing to perform well in real-world settings.  
In terms of transparency in shared task organisation, 
\citet{knowles-etal-2020-nrc} provide a ``Statement on Avoiding Conflicts of Interest" that describes the separation of roles with respect to organisers and participants.

In a broader context, it should be noted that there have been general steps towards addressing some of the issues raised by \citet{parraescartin-EtAl:2017:EthNLP} and in our survey here. Along with shared tasks/workshops that focus on the need for ensuring reproducibility in reported studies (e.g. \citet{Branco4REAL:2016}, \citet{Branco4REAL:2018} and \citet{branco-EtAl:2020:LREC2}), encouragingly, there is an increasing awareness around the need for reproducibility in the field (e.g. \citet{dakota-kubler-2017-towards}, \citet{wieling-etal-2018-squib} and  \citet{cohen-etal-2018}), while some of the major conferences (e.g. ACL-IJCNLP\footnote{\url{https://2021.aclweb.org/calls/papers/\#reproducibility-criteria}}) now encourage authors to submit supplementary material to facilitate others in reproducing their results. Additionally, in terms of responsible dataset curation and management, ACL-IJCNLP organisers request that both reviewers and authors consider the potential ethical concerns relating to the sourcing and management of the data used in their studies.\footnote{\url{https://2021.aclweb.org/ethics/Ethics-review-questions/}}

We have also seen the introduction of workshops that encourage the reporting and analysis of negative results (e.g. the First Workshop on Insights from Negative Results in NLP \cite{insights-2020} or the NeurIPS 2020 Workshop ``I Can't Believe It's Not Better!''(ICBINB)\footnote{\url{https://i-cant-believe-its-not-better.github.io}}). That said, the field as a whole would also benefit from a journal that encourages the publication of negative results. 
%such as that established in the field of Biomedical Sciences (Journal of Negative Results in Biomedicine).\footnote{\url{https://jnrbm.biomedcentral.com}}
In fact there was an attempt made towards this back in 2008, the Journal of Interesting Negative Results in Natural Language Processing and Machine Translation\footnote{\url{http://jinr.site.uottawa.ca/}} which seems to not have gathered the attention from our community that it could have.

In the next section (\ref{sec:ST_checklist}), we provide a suggested checklist for future shared tasks (and extending to benchmarking efforts, grand challenges, evaluation campaigns etc.) based on the collation of both the respondents' answers to our direct questions along with their comments. While this checklist is of course based on current perceptions of shared tasks, we note that shared tasks (and the field as a whole) are evolving continuously and that such a checklist should be modified over time as necessary.
\section{Shared Task Organisation Checklist}
\label{sec:ST_checklist}

%As a direct result of this survey and the feedback we received,
Here we provide a common framework which (i) shared tasks organisers can use as a reference in defining their task and (ii) will highlight to participants and annotators the factors that can vary across shared tasks, some of which they may need to seek clarification on. Therefore, the purpose is not to formalise shared tasks as a whole but instead to prompt organisers to consider the wide range of positive and negative sides of shared tasks -- from annotators', participants' and organisers' perspectives.

%\todo{not sure this fits}The suggestion is that organisers can provide their relevant information not only in their promotional announcements and on their website, but also in their findings paper to ensure a sustainable artefact is available.

The idea behind the following checklist is for it to become a guideline for encouraging common standards
%high levels of transparency 
in the participation and organisation of shared tasks. If organisers and participants alike opt to use the checklist when announcing the shared task and/or submitting systems to them, the entire community should benefit, as greater levels of transparency would be achieved, and potentially the issues discussed in the previous sections \ref{ssec:participationResults} and \ref{ssec:organisationResults} would be mitigated against. To allow for a clear interpretation as to why each issue in the checklist should be considered, we have categorised them according to various factors that could affect their success.
%If specific details are required, then they should accompany the statement (e.g. but their system will not be ranked, only in the case of open data use)

\paragraph{Transparency}

\begin{itemize}
\item Can organisers participate?
\item Can annotators participate?
\item Can evaluators participate?
\item Is there a global mailing list to keep participants in the loop with responses to any questions or changes?
\item Will timelines be provided from the announcement of the shared task and will they be publicly available?
 \item Are timelines realistic and will unforeseen changes be communicated promptly?
\item What is the type of shared task (open/closed/both)?
\item Who is the invited participant audience (Research institution teams only? Industry teams also? Will there be separate tracks?)
\item Is the data collection process documented? Did the data exist already from other studies or are they being curated or annotated specifically for this task?
\item Have the licensing conditions for sharing data been verified?
\item Where relevant, have steps been taken to safeguard privacy rights or sensitive data (e.g. anonymisation)?
\item If manually annotated by crowd workers, were annotators fairly compensated?
\item Are there clear annotation guidelines for new datasets being used in the shared task and are they publicly available?
\item Is there consistency in annotation and format across multiple datasets?
\item Are details of any changes in data formats or annotation compared to previous tasks clearly communicated?
\item If an annotation tool is provided, are there one or two organisers representing technical support points of contact, should any issues be encountered?
\item Is there a (documented and accessible) quality control step before releasing datasets for the task?
\item How will systems be evaluated (remote VMs, web-based interfaces, APIs, etc.)?
\end{itemize}

\paragraph{Reporting and Replicability}

%\begin{itemize}
%\item If applicable, the requirement of a system description paper from each team is clearly stated.
%\item A template for the system description is provided to encourage consistency across participants' reporting.
%\item Complex systems are allowed to submit longer papers to allow for detailing their complexity for reproducibility purposes.
%\item The system description papers can be accompanied by supplementary materials.
%\item Those ranking below a certain threshold are encouraged to submit system descriptions to learn from negative results.
%\item Participants are encouraged to also provide an error analysis alongside negative results.
%\item The provision of an overview paper and the level of detail that will be reported is clearly stated in the website.
%\item It is clearly stated who will be named authors on publications initiated by the shared task organisers (participants, annotators, evaluators, etc.).
%\item If specific details are required, a clear statement has been released in the shared task website (e.g. participants that do not submit a description paper will not be ranked; participants using additional proprietary data will not be ranked, etc.).
%\end{itemize}

\begin{itemize}
\item Is the requirement of a system description paper from each team made clear?
\item Is there a template for the system description to encourage consistency across participants' reporting?
\item Can shorter description papers from those ranking below a certain threshold be submitted?
\item Can longer description papers for complex systems be submitted?
\item Can system description papers be accompanied by supplementary materials?
\item Are participants encouraged to also provide an error analysis alongside negative results? 
\item Is information clearly provided on whether or not an overview paper will be produced and how much detail it will contain?
\item Is it clear to participants who should be named as authors on each publication (participants, annotators, etc.)?
%\begin{itemize}
%\item The shared task requires that participants release their code (including a detailed README) after the shared task.
%\item The shared task requires that participants release their data after the shared task (where copyright/licensing allows).
%\item The shared task requires that participants release a detailed list of all data used in their system submissions.
%\end{itemize}
\item Are participants required to release their code (including a detailed README) after the shared task?
\item Are participants required to release their data after the shared task (where copyright/licensing allows)?
\item Are participants required to release a detailed list of all data used in their system submissions?
\item Are other specific details made clear on the shared task website (e.g. participants that do not submit a description paper will not be ranked; participants using additional proprietary data will not be ranked, etc.)?
\end{itemize}

\paragraph{System Ranking}

%\begin{itemize}
%\item Participants are encouraged to share negative results.
%\item Participants can choose not to be ranked if so preferred.
%\item Participants who rank low (below a chosen threshold) can choose to be anonymised.
%\end{itemize}

\begin{itemize}
\item Are participants encouraged to share negative results and can they choose not to be ranked if so preferred?
\item Can participants who rank low (below a chosen threshold) choose to be anonymised?
\item Is it possible to report only the top N-ranked system scores in order to avoid withdrawal or fear of negative results?
\end{itemize}

\paragraph{Metrics}

%\begin{itemize}
%\item The evaluation metrics are clearly stated when the shared task is announced.
%\item All metrics that will be used for ranking systems are clearly stated.
%\item Evaluation scripts will be shared after the shared task (or, if applicable, are already available).
%\end{itemize}

\begin{itemize}
\item Are evaluation metrics clearly stated when the shared task is announced?
\item Are all metrics to be used for ranking systems clearly stated?
\item Will evaluation scripts will be shared after the shared task (or, if applicable, are they already available)?
\end{itemize}

The above lists are provided as checklists for organisers and participants. Our overall suggestion is that organisers can provide their relevant information not only in their promotional announcements and on their website, but also in their findings paper to ensure a sustainable artefact of the organisation of each task will be available in the future.
\section{Conclusion}
\label{sec:conclusion}

This paper has summarised the results of a survey carried out to ascertain the degree to which the current organisation of and participation in shared tasks pose ethical risks to the NLP community. The survey elicited a broad and informative response from the NLP field with respect to respondents' past experiences relating to shared tasks (either as organisers, annotators or participants). The survey questions were designed to and succeeded in gathering information regarding (i) best practices used in specific shared tasks that could be extrapolated to new ones, (ii) the type of information that is usually available to participants before, during and after the shared task, (iii) potential ethical concerns encountered in the past and how they were tackled, (iv) other causes for concern from the NLP community and (v) positive experiences that we should aim to replicate.

The survey found that, according to 170 respondents, while shared tasks are generally celebrated as an important factor in advancing the field of NLP, there is a wide range of issues that both participants and organisers would like to see improvements on. In many cases, opinions were mostly unanimous in terms of the need for more transparency in the organisation of shared tasks. This includes resource sharing for the sake of replicability, clarity with respect to reporting levels, and transparency around evaluation metrics, management of data and levels of participation. We presented the results in a manner which provides further insight as to the types and level of experience of shared tasks that informed the responses.

The reader should be aware that the purpose of carrying out this survey was to avoid a scenario whereby a small number of authors would make recommendations for the improvement of shared tasks based solely on their own limited experiences. As we have seen, the number and type of shared tasks are too wide and varied and hence there is no ''one-size-fits-all" formula that would ensure that, going forward, no shared tasks could avoid encountering ethical risks. However, our survey results clearly showed that our community would benefit from clearer information about shared tasks, as well as a framework that could help organisers reflect on factors that need to be considered when organising a shared task.

%In Section \ref{sec:ST_checklist} 
We have proposed a Shared Task Organisation Checklist as that guiding framework. This checklist is a guideline of suggestions on how to successfully and transparently run shared tasks in NLP, based on feedback from the community itself. 
\bibliographystyle{spbasic}      % basic style, author-year citations

\bibliography{compling.bib}   % name your BibTeX data base

%\begin{thebibliography}{}
% etc
%\end{thebibliography}

\end{document}